\documentclass[journal]{IEEEtran}         

\usepackage[dvipsnames]{xcolor}
\usepackage{amsthm} 
\usepackage{amsfonts}
\usepackage{amsmath}
\newtheorem{theorem}{Theorem}
\theoremstyle{plain}

\usepackage{algorithm}
\usepackage[noend]{algpseudocode}

\usepackage{graphicx}
\usepackage{todonotes}
\usepackage{amssymb}
\usepackage[caption=false]{subfig}
\usepackage[capitalise]{cleveref}

\usepackage{subcaption}
\usepackage{cite}

\newtheorem{problem}{Problem}
\title{Homotopic information gain for \\sparse active target tracking}
\author{
    Jennifer Wakulicz,
    Ki Myung Brian Lee,
    Teresa Vidal-Calleja,
    Robert Fitch
\thanks{
    Jennifer Wakulicz is with the Australian Centre for Robotics, School of Aerospace, Mechanical and Mechatronic Engineering, University of Sydney, Camperdown, 2006, NSW, Australia (email: jennifer.wakulicz@sydney.edu.au).}%
\thanks{Ki Myung Brian Lee is with the Department of Electrical and Computer
Engineering, UC San Diego, 9500 Gilman Drive, La Jolla, CA 92093-0411,
USA (email: kmblee@ucsd.edu).}%
\thanks{
    Robert Fitch and Teresa Vidal-Calleja are with the School of Mechanical and Mechatronics Engineering, University of Technology Sydney, Ultimo, 2007, NSW, Australia 
    (email: \{rfitch, Teresa.Vidal-Calleja\}@uts.edu.au).
}
}

\begin{document}

\maketitle

\begin{abstract}
The problem of planning sensing trajectories for a mobile robot to collect observations of a target and predict its future trajectory is known as active target tracking. Enabled by probabilistic motion models, one may solve this problem by exploring the belief space of all trajectory predictions given future sensing actions to maximise information gain. However, for multi-modal motion models the notion of information gain is often ill-defined. This paper proposes a planning approach designed around maximising information regarding the target's homotopy class, or high-level motion. We introduce homotopic information gain, a measure of the expected high-level trajectory information given by a measurement. We show that homotopic information gain is a lower bound for metric or low-level information gain, and is as sparsely distributed in the environment as obstacles are. Planning sensing trajectories to maximise homotopic information results in highly accurate trajectory estimates with fewer measurements than a metric information approach, as supported by our empirical evaluation on real and simulated pedestrian data.
\end{abstract}


\section{Introduction}
Active target tracking is the task of choosing informative sensing actions or policies regarding a time-evolving target of interest such that an estimate of its state can be accurately maintained~\cite{pmlr-v211-yang23a,zhou_resilient_2019,zhou_active_2019}. When the state under estimation is the target's full trajectory, active target tracking has important applications in robotics problems such as surveillance, search and rescue and human-robot or multi-robot coordination. 

Solving the active target tracking problem requires addressing two challenges: trajectory prediction -- to predict where the target likely will be in the future, and information gathering -- to decide at which location and time to observe the target to maximise understanding of its trajectory. Trajectory prediction models are well studied, capturing a broad range of dynamics from linear with unknown components~\cite{Gillijns2007A}, to more general non-linear dynamics~\cite{li_attentional-gcnn_2020,zhi_kernel_2020}, even incorporating contextual constraints into learned models~\cite{kiss_constrained_2022,eiffert_probabilistic_2020,wakulicz_topological_2023}. Yet, from the sensing robot's perspective, no matter how accurate a motion model is, the target's future trajectory is still stochastic in nature. Planning where to visit to observe the target is therefore a problem marred with uncertainty. 
If the predictive motion model provides trajectory estimates represented as probability distributions, referred to as beliefs, the sensing robot can reason under this uncertainty directly by planning in the space of possible future outcomes, known as the belief space. 

To plan informative sensing trajectories in belief space the robot must search for sensing actions that lead to beliefs of minimal uncertainty. However, exploring belief spaces is notoriously challenging. They are of high dimension and transitions between beliefs given sensing actions are stochastic~\cite{kurniawati_partially_2022}. The value of a sensing action can therefore typically only be estimated by estimating the \textit{expected} information gain it provides. This requires evaluating information-theoretic cost functions that can be computationally expensive or even intractable. For example, multi-modal or multi-expert beliefs such as those in~\cite{eiffert_probabilistic_2020,zhi_kernel_2020,wakulicz_topological_2023}, while producing more accurate predictions than their single-mode counterparts, often have no analytic form for information gain. For multi-modal beliefs modeled by Gaussian mixture models (GMMs), approximations and upper bounds for information gain have been derived~\cite{hershey_approximating_2007,kolchinsky_estimating_2017}. However, these approximations often require pairwise calculations between components, becoming cumbersome to evaluate for large models.

\begin{figure}[t!]
    \centering
         \includegraphics[width=0.48\columnwidth]{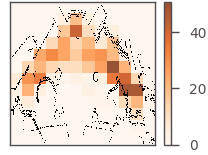} \hfill
         \includegraphics[width=0.48\columnwidth]{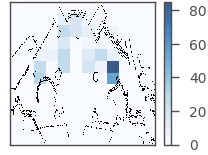}
    \caption{Heatmaps showing visitation frequency at sensing locations for the active target tracking problem on the ATC dataset. A conventional tracking approach (left) is compared to the sparse homotopic approach proposed in this paper (right). With our proposed approach, the sensing robot is only concerned with how a target manoeuvres around obstacles, reducing planning and sensing resources while still producing accurate estimates of the target trajectory. \label{fig:spatial_heatmap}}
\end{figure}


In this paper we propose to solve the active target tracking problem in the high-level belief space admitted by the hierarchical GMM developed in~\cite{wakulicz_topological_2023}. Here the authors present a model that predicts the \textit{homotopy class} of a given agent's trajectory. The homotopy class of a trajectory captures its high-level motion, i.e., whether it turns left or right at an obstacle. The probabilistic homotopy prediction, which we refer to as a homotopic belief, is then fed to a GMM to produce low-level trajectory predictions. Thus, 
we propose to plan sensing trajectories in the space of homotopic beliefs. 

As the homotopy class of a trajectory is a discrete random variable with a countable set of outcomes, the associated belief space is discrete and low-dimensional. Moreover, since 
the homotopy class of a target is determined by how it manoeuvres around obstacles, information in the homotopic belief space is as sparsely distributed as obstacles are. We show that planning over this sparse, discrete latent belief space therefore allows for efficient active target tracking. Furthermore, we derive a connection between homotopic information gain and metric information gain guaranteeing that planning in the homotopic belief space provides high-quality metric or low-level information, maintaining accurate trajectory estimates despite the computational advantage. Our sparse and accurate active target tracking approach thus opens exciting avenues for overcoming resource constraints in multi-target or multi-task problems.

The contributions of this paper are summarised as follows.
\begin{itemize}
    \item The proposal and construction of homotopic information gain as a metric for planning informative sensing trajectories for active target tracking.
    \item Theoretical analysis demonstrating homotopic information gain is a lower bound for metric information gain.
    \item An online planning framework for solving the active target tracking problem using our proposed homotopic information gain.
    \item Evaluation and testing of the proposed information gain and planning framework on simulated and real pedestrian datasets with varying complexity.
\end{itemize}
Moreover, code will be open-sourced.



\section{Related Work}

\subsection{Homotopy classes in robotics}\label{sec:bg:homotopy_theory}
In robotics, it is often useful to classify a mobile agent's motion into high-level categories. For example, a description of an external agent turning left or right around an obstacle can be enough information for an autonomous robot to plan a non-colliding path. In topology, the \emph{homotopy class} of an agent's path captures precisely this information.
Two paths $\tau_1$, $\tau_2$ in a topological space $\mathcal{D}$ with common start and ending points belong to the same homotopy class if there exists a continuous transformation or deformation from one to the other~\cite{hatcher_algebraic_2002}. For example, paths $\tau_1$ and $\tau_2$ in~\cref{fig:homotopy} are homotopic -- there is a continuous transformation between them illustrated with dotted paths. However, $\tau_1$ and $\tau_2$ are not homotopic to $\tau_3$ as any deformation requires moving through the obstacle $\mathcal{O}$ and breaking continuity.
Obstacles therefore induce non-trivial homotopy classes, and split the possible ways an agent can move through an environment into abstract motion classes. To illustrate this, in~\cref{fig:homotopy:class} the obstacle $\mathcal{O}$ induces homotopy classes of trajectories that move `above', `below', and wind around the obstacle. Such abstraction of the high-level motions available to an agent is a powerful tool for navigation, planning and prediction tasks often encountered in robotics.

Homotopy classes provide the ability to reason over a diverse set of paths when performing robotic path planning. Maintaining diversity in path planning can be beneficial in dynamic applications, allowing the robot to switch between paths rather than re-plan new ones when mission definitions are changed. This becomes particularly useful when path planning for autonomous navigation in environments where other dynamic agents are present. In~\cite{de_groot_topology-driven_2025,kolur_online_2019}, an optimal trajectory from each homotopy class is maintained online to plan fast and decisive collision-free navigation through dynamic spaces. Alternatively, with knowledge of the diverse set of options, one can constrain the planning problem to a specific homotopy class or exclude specific classes. For example, in~\cite{bhattacharya_search-based_2010} a homotopy-aware A* variant is developed, allowing one to find the optimal path for any chosen homotopy class through discretised space. Similar homotopy-aware extensions of classical sampling-based planners PRM and RRT have also been developed to this end in~\cite{novosad_ctopprm_2023,yi_homotopy-aware_2016,sakcak_convex_hsig}.

Abstracting trajectories into high-level motions lends itself naturally to trajectory prediction, a task relevant to robotics problems such as navigation, search and rescue and human-robot interaction. Predictive models that provide multiple predictions of an agent's future path with varying likelihoods such as~\cite{zhi_kernel_2020,eiffert_probabilistic_2020} capture the \emph{multi-modal} nature of an agent's possible future paths, showing greater prediction accuracy over single-mode approaches~\cite{wiest_probabilistic_2012,Ivanovic_2019_ICCV,zyner_naturalistic_2020}. Homotopy classes provide a natural language to describe the multi-modal nature of an agent's future trajectories. In~\cite{pokorny_topological_2016,frederico_carvalho_long-term_2019}, a homotopy-informed clustering method is developed to identify high-level motion primitives through environments, represented as vector flow fields that are then used for prediction. In~\cite{wakulicz_topological_2023} a homotopic GMM is proposed, where each GMM component represents a homotopy class in the training data and weights are fused with homotopic predictions for improved pedestrian prediction.

\subsection{Informative path planning for active target tracking}

Active target tracking utilises predictive motion models to plan sensing trajectories that improve a robot's estimate of the target's location or trajectory. Where state estimates are formulated and maintained probabilistically, approaches to active target tracking commonly involve planning in belief space. Probabilistic estimates are referred to as beliefs, giving not just the state estimate but also the uncertainty in the estimate. Then, rather than planning sensing actions in physical space, one can plan to `visit' beliefs with minimal uncertainty.

For targets whose motion models can be sufficiently described by linear dynamics with additive Gaussian noise, a Kalman filter provides the tools required for belief space planning. Here, the belief space is Gaussian -- all estimates are Gaussian distributions, defined solely by the most likely state estimate and the corresponding covariance given a sensing location. To predict a belief at a future sensing location, the Kalman update equations are readily available. The ease of navigating Gaussian belief spaces make Gaussian and most-likely estimate assumptions widely popular in active target tracking literature~\cite{Pappas2009,vitus_closed_loop_2011,Tomlin2012,atanasov_information_2014,tzoumas_sensor_2016,Schlotfeldt2018,kantaros2019}. Recovering Gaussian belief space planning in scenarios with non-linear target and sensing dynamics is made possible via the extended Kalman filter as in~\cite{ke_zhou_multirobot_2011,Tokekar2016,zahroof_multi-robot_2023}. In cases where noise is also non-Gaussian, beliefs can instead be approximated via sampling using particle filters~\cite{ryan_particle_2010,tokekar_multi-target_2014,pmlr-v119-fischer20a}.

Multi-modal or multi-hypothesis models improve the accuracy of trajectory prediction and are thus powerful tools for active target tracking. However, they can be overwhelming in belief space planning scenarios. With each measurement and belief update, the number of possible motion hypotheses grows. There has been plentiful work on addressing this combinatorial expansion for beliefs represented by GMMs, involving merging or pruning components at each fusion step~\cite{vo_gaussian_2006,ruijie_he_efficient_2010,kim_multiple_2015,zheng_gaussian_2024,billard_nonmyopic_2023}. Simplistic solutions fix the number of possible motion hypotheses to a constant number. Then, a component-wise belief update can be performed via standard Gaussian conditioning as in~\cite{wiest_probabilistic_2012,wakulicz_topological_2023}.

With manageable multi-hypothesis belief spaces, optimal sensing trajectories for active target tracking can be posed as those that maximally reduce the expected uncertainty or entropy in the resulting belief, as in single-hypothesis cases~\cite{atanasov_information_2014,wakulicz_active_2021}. However, for multi-hypothesis beliefs such as GMMs, the notion of entropy is not analytically defined~\cite{carreira-perpinan_mode-finding_2000} and so approximations must be made. There are many proposed approximations~\cite{tian_lan_estimating_2006,kolchinsky_estimating_2017}. In~\cite{kolchinsky_estimating_2017}, for example, bounds for mixture entropy are given via calculation of pairwise distances between components. Still, for large GMMs with many components a pairwise calculation can become computationally expensive. In this paper, we instead propose to plan sensing trajectories that reduce the expected \emph{homotopic uncertainty} to hone in on a single, true hypothesis of a pedestrian whose trajectory is modeled by the homotopic GMM proposed in~\cite{wakulicz_topological_2023}. 


\section{Background and problem formulation}

 \begin{figure}[t!]
    \centering
    \subfloat[Paths of equal ($\tau_1$, $\tau_2$) and differing ($\tau_3$) homotopy classes.]{\includegraphics[width=0.37\linewidth]{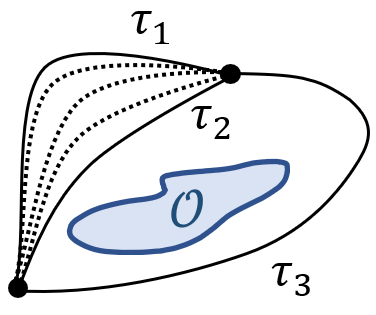}\label{fig:homotopy:class}}\hfill
    \subfloat[Construction of $h(\gamma) = (1,2,-2)$ by following $\gamma$ as it crosses rays emanating from obstacles.]{\includegraphics[width=0.55\linewidth]{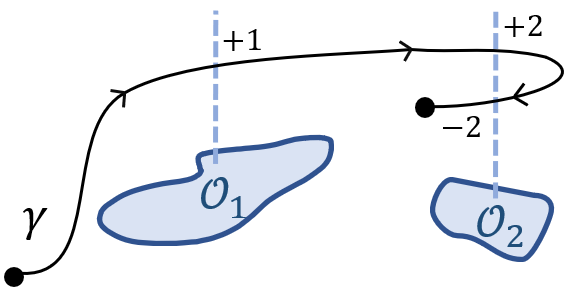}\label{fig:homotopy:hsig}}
    \caption{Illustrations of homotopy classes and $h$-signature calculation.~\cite{wakulicz_topological_2023}}\label{fig:homotopy}
    \vspace{-1.5em}
\end{figure}

\subsection{The $h$-signature as a homotopy invariant}\label{sec:bg:invariants}
Homotopy invariants are signatures of paths that remain unchanged for any path within the same homotopy class. That is, for two paths $\tau_1$ and $\tau_2$, the homotopy invariants $h(\tau_1)$, $h(\tau_2)$ of each path are equal if and only if $\tau_1$ and $\tau_2$ are homotopic. Such invariants are useful in robotics, as they avoid the requirement to check for existence of continuous transforms between paths. A commonly used homotopy invariant is the \emph{$h$-signature}~\cite{hatcher_algebraic_2002} for the ease of its calculation. The $h$-signature can be thought of as a `word' constructed by following the path across obstacles. First, rays are drawn from a point within each obstacle to the boundary of the environment, ensuring no rays intersect one another. Each ray is given an arbitrary but unique `letter' $n$, in this paper we choose letters corresponding to the enumeration of the obstacles. Then, by tracing a given path the word may be constructed, appending the letter `$n$' of a ray if the path crosses from left to right, and `$-n$' if the path crosses from right to left. For example, in~\cref{fig:homotopy:hsig}, the path $\gamma$ crosses the first ray from left to right, then the second, before crossing the second ray again from right to left. The $h$-signature of $\gamma$ is therefore `$(+1,+2,-2)$'. This can be further reduced by canceling adjacent identical letters with opposite sign, i.e. the reduced $h$-signature of $\gamma$ is simply $(+1)$. 

The simple computation of $h$-signatures allows one to uniquely identify the high-level motion or homotopy class of any trajectories sharing common start and end points. However, in robotics we must often reason with trajectories with varying endpoints. To overcome this limitation, the quotient map of a space is often taken, as in~\cite{mccammon_topological_2021,pokorny_topological_2016}. The quotient map takes all boundary points of an environment to a common quotient point while preserving the topology of the space. Thus, any trajectories starting and ending on the boundary of the environment effectively begin and end on the quotient point and can be compared homotopy-wise with $h$-signatures. Here, paths transform into `loops' beginning and ending at the quotient point, and the homotopy class captures the obstacles enclosed in the loop and the way in which the loop winds around them.

\subsection{The homotopic belief space}
In~\cite{wakulicz_topological_2023}, the partial $h$-signature $\rho$ is introduced to capture the high-level motion of an incomplete trajectory. It is calculated following the same process described in~\cref{sec:bg:invariants}, only on partially completed trajectories rather than full ones. As partial trajectories do not end on the quotient point, they are not true homotopy invariants, and are only used as a predictor for the full $h$-signature of a path. Furthermore, in an online context the partial $h$-signature cannot be directly observed but must instead be inferred from a sequence of measurements of the partially completed trajectory. The partial and full $h$-signatures are therefore treated as random variables.

With an inferred partial $h$-signature, a probability distribution over the possible future full $h$-signatures can be constructed using a probabilistic sequence completion framework such as a Variable Order Markov Process (VOMP). In particular, the conditional probability of a particular outcome $h_*$ is given by
\begin{equation}\label{eq:homotopy:naive_conditioning}
    p(h=h_* \mid \rho) = \frac{p(h_*)}{\sum_{h' \in \mathcal{H}(\rho)\setminus h_*} p(h') }, 
\end{equation}
where $\mathcal{H}(\rho)$ is the set of full $h$-signatures compatible with $\rho$ and $p(h_*)$, $p(h')$ are given by the VOMP. For a given partial $h$-signature $\rho$, compatible full $h$-signatures are those whose prefix is $\rho$, i.e., $\mathcal{H}(\rho) = \{h \mid \exists \rho', h = \rho\rho^{\prime}\}$. We restrict the notion of compatibility to apply only to unreduced $h$-signatures. Otherwise the process of reduction may be exploited to render all $h$-signatures compatible with a partial one. To improve computation time and prevent hallucination of the VOMP, we diverge from the methodology in~\cite{wakulicz_topological_2023} and further restrict $\mathcal{H}(\rho)$ to the set of $h$-signatures observed in the training data following the assumption that no out-of-distribution trajectories are present in the test set.

Since $p(h\mid \rho)$ is a probabilistic estimate of the homotopy class of a target's trajectory, we henceforth refer to this distribution as the \emph{homotopic belief}. In this paper we propose planning over the \emph{homotopic belief space} to solve the active target tracking problem, formalised in the next section. 

\subsection{Problem formulation}\label{sec:bg:prob_form}

Consider an agent traversing through a $2D$ environment $\mathcal{D}$ containing $n$ obstacles $O = \{\mathcal{O}_1,\ldots,\mathcal{O}_{n}\}$.
The agent's path $\mathbf{Y}$ is a sequence of positions $\mathbf{y}_t \in \mathbb{R}^2$ at a sequence of time steps $t\in[0,T-1]$. That is, $\mathbf{Y} = \{ \mathbf{y}_{0}, \ldots \mathbf{y}_{T-1} \}$.
We assume that the agent begins and ends its path on the boundary $\delta\mathcal{D}$ of the environment.

A mobile sensing robot with state $\mathbf{x}_t\in\mathbb{R}^{2}$ at time $t$ is tasked with locating and measuring the agent's state to build an accurate estimate of its full trajectory $\mathbf{Y}$, both elapsed and future. The robot takes noisy measurements $\mathbf{z}_t$ of the agent at time $t$ following the sensor observation model
\begin{equation}\label{eq:prob:msmt_model}
    \mathbf{z}_t = \mathbf{y}_t + \mathbf{v}_t(\mathbf{x}_t),
\end{equation}
where $\mathbf{v}_{t}(\mathbf{x}_t)\sim \mathcal{N}(\mathbf{0},\mathbf{R}(\mathbf{x}_t))$ is additive measurement noise. To aid in reconstruction and prediction of $\mathbf{Y}$, a dataset of $K$ historical, fully observed trajectories $\mathcal{Y} = \{\mathbf{Y}_{\text{obs}}^1,\ldots,\mathbf{Y}_{\text{obs}}^K\}$ is assumed to be available for learning of a predictive motion model, as is full knowledge of obstacle locations. To study and compare homotopy classes of the historical trajectories, we must assume them to begin and end on boundary locations.

More formally, the sensing robot is tasked with the following information gathering problem.
\begin{problem}\label{prob:active_perception}
    Given a planning horizon $\tau < \infty$ and an initial belief over $\mathbf{Y}$, choose a sequence of sensing locations $\sigma = \{\mathbf{x}_1, \ldots, \mathbf{x}_{\tau}\}$ for the robot that maximises the information gain of the agent's trajectory $\mathbf{Y}$ given the measurement set, or
    \begin{equation}
        \begin{array}{ll}
            \sigma^* = \arg\max\limits_{\sigma\in \mathcal{X}^{\tau}}& \mathcal{I}(\mathbf{Y}; \mathbf{z}_{1:\tau} \mid \sigma) 
        \end{array}
    \end{equation}
    where $\mathbf{z}_{1:\tau}=\{\mathbf{z}_1,\ldots,\mathbf{z}_\tau\}$ is the set of measurements taken up to time $\tau$, and $\mathcal{X}^\tau$ is the set of all possible sensing sequences.
\end{problem}

\section{Homotopic information gain for active target tracking}\label{sec:homotopic_info}
Solving~\cref{prob:active_perception} requires calculating the expected information gain at future sensing locations. This is typically captured by calculating the Kullback-Liebler (KL) divergence between the trajectory belief before and after measurement, that is, between the prior and posterior probability distributions estimating the trajectory. The trajectory belief can be modeled and maintained by any probabilistic model.

\subsection{Gaussian mixture models for trajectory belief modelling}
To model low-level trajectories we adopt the homotopic GMM proposed in~\cite{wakulicz_topological_2023}, where each Gaussian component models the motion of a unique homotopy class. The homotopic GMM gives probability distribution $ p(\mathbf{Y} \mid h)$ over the full trajectory $\mathbf{Y} = \{ \mathbf{y}_{0}, \ldots, \mathbf{y}_{T-1} \}$ given $h$.
To train the model training trajectories from historical dataset $\mathcal{Y}$ are clustered by $h$-signature $h'$. Then, a Gaussian is fit to each cluster giving the following GMM
\begin{equation}\label{eq:homotopy:homotopic_gmm}
    p(\mathbf{Y} \mid h) = \sum_{h'} w^{h'} \mathcal{N}(\boldsymbol{\mu}^{h'}, \Sigma^{h'}),
\end{equation}
where $w^{h'}$, $\boldsymbol{\mu}^{h'}\in\mathbb{R}^{2T}$ and $\Sigma^{h'}\in\mathbb{R}^{2T\times2T}$ are the weight, mean vector and covariance respectively of homotopy class $h'$. For cases where there is multi-modal low-level behaviour within one homotopy class (exemplified in~\cref{fig:thor_data_hsig}), $N_C$ components can be fit. This further increases the concentration of covering Gaussians, better constraining each Gaussian to a homotopy class. Where $N_C \neq 0$, the above sum becomes a double sum over both $h$ and $c\in\{1,\ldots,N_C\}$.




The GMM can be used to derive a probabilistic prediction of the trajectory given the measurements $\mathbf{z}_{1:t}$ and corresponding partial $h$-signature $\rho_t$. As in~\cite{wakulicz_topological_2023}, we treat the partial $h$-signature as a random variable conditionally independent of $\mathbf{Y}$ given the full $h$-signature $h$. Then, we can easily condition~\cref{eq:homotopy:homotopic_gmm} on the partial $h$-signature as a weighted sum
\begin{equation}
p(\mathbf{Y} \mid \rho_{t}) = \sum_{h'\in\mathcal{H}(\rho_t)} p(\mathbf{Y} \mid h') p(h' \mid \rho_{t}).
\end{equation}
This amounts to simply scaling the weights of the GMM by the homotopic belief $p(h\mid \rho_{t})$ maintained by the VOMP.
Then, to condition on the observed low-level measurements $\mathbf{z}_{1:t}$ we use standard Gaussian conditioning~\cite[Section~8.1.3]{matrixcookbook} to find the posterior mean $\hat{\boldsymbol{\mu}}$ and covariance $\hat{\Sigma}$ of each component, giving the posterior GMM
\begin{equation}\label{eq:homotopy:conditioning}
    p(\mathbf{Y} \mid \rho_{t}, \mathbf{z}_{1:t}) = \sum_{h'\in\mathcal{H}(\rho_t)} \hat{w}^{h'} \mathcal{N}(\hat{\boldsymbol{\mu}}^{h'}, \hat{\Sigma}^{h'} ),
\end{equation}
with conditional weights
\begin{equation}
    \hat{w}^{h'} \propto w^{h'} p(h' \mid \rho_{t}) \mathcal{N}(\boldsymbol{\mu}^{h'}, \Sigma^{h'} + \sigma_{Z}^2 I), \label{eq:homotopy:conditioning_weights}
\end{equation}
followed by re-normalisation.

With the homotopic GMM model in place the expected information gain given a sensing location can be used to plan sensing trajectories. 
However, there is no analytic form for information gain where the belief is represented by a GMM~\cite{carreira-perpinan_mode-finding_2000}. Approximations and upper bounds for these information gain measures exist~\cite{tian_lan_estimating_2006,kolchinsky_estimating_2017}, requiring pairwise calculations between GMM components. For large search spaces and GMMs, the required computation time for these approximations can therefore prohibit solving~\cref{prob:active_perception} in a feasible time. Rather than facing this challenge, we propose to instead plan over the homotopic belief space to maximise \textit{homotopic} information gain.

\subsection{Homotopic information gain}
We formulate and solve the active target tracking problem by planning sensing actions that capture homotopic information. That is, we propose the following problem.
\begin{problem}\label{prob:homotopic_tracking}
    Given a planning horizon $\tau < \infty$ and an initial belief $h$ over the $h$-signature of a trajectory $\mathbf{Y}$, choose a sequence of sensing locations $\sigma = \{\mathbf{x}_1, \ldots, \mathbf{x}_{\tau}\}$ for the sensing robot that maximises the information gain of the agent's homotopy glass $h$ given the measurement set, or
    \begin{equation}
        \begin{array}{ll}
            \sigma^* = \arg\max\limits_{\sigma\in \mathcal{X}^{\tau}}& \mathcal{I}(h; \rho_{\tau}, \mathbf{z}_{1:\tau} \mid \sigma) 
        \end{array}
    \end{equation}
    where $\mathbf{z}_{1:\tau}=\{\mathbf{z}_1,\ldots,\mathbf{z}_\tau\}$ is the set of measurements taken up to time $\tau$, $\rho_\tau$ is the partial $h$-signature inferred from the measurement set, and $\mathcal{X}^\tau$ is the set of all possible sensing sequences.
    
\end{problem}

By solving~\cref{prob:homotopic_tracking} rather than~\cref{prob:active_perception}, we aim to plan sensing trajectories that reduce uncertainty not in the target's \textit{low-level} trajectory, but in its more abstract \textit{high-level} motion. This encodes some real-world intuition into our sensing policies: to track a target through a cluttered environment one might wait at a location where the target must make a high-level decision such as whether to move left or right around an obstacle. 


The information about a random variable gained from a measurement is captured by the KL divergence between the belief of the random variable before and after the measurement is taken. Here, the belief of the homotopy class of a path only changes when a measurement is taken that changes the inferred partial $h$-signature, that is, when a ray crossing is observed. The current partial $h$-signature can be inferred directly by tracing a rough path constructed from the measurement set. However, if measurements are very sparse, this can lead to inaccurate partial $h$-signatures. For a more rigorous approach, the homotopic belief $b_t$ at time $t$ can be marginalised over possible partial $h$-signatures given the measurement set up to time $t$, that is, $b_t = \sum_{c}p(\rho_t \mid c)p(h \mid \rho_t)$, where $p(\rho_t\mid c)$ is the weight of the posterior Gaussian component $c$ from which $\rho_t$ is extracted (\cref{eq:homotopy:conditioning_weights}). 

With the partial $h$-signature inferred, the homotopic information gain is then the KL divergence between the current homotopic belief $b_t$ and the new homotopic belief $b_{t+1}$. 
\begin{equation}\label{eq:homotopic_KL}
    D_{\textrm{KL}}(b_t,b_{t+1}) = \sum_{h'\in\mathcal{H}(\rho_{t+1})} p(h'\mid\rho_{t})\log\left(\frac{p(h'\mid\rho_{t})}{ p(h'\mid\rho_{t+1})}\right).
\end{equation}

Realistically, the future partial $h$-signature inferred from the measurement set $\mathbf{z}_{1:t+1}$ at a future time $t+1$ given future sensing location $\mathbf{x}_{t+1}$ is not known due to sensing and modeling uncertainty. We must therefore calculate the expected information gain given a sensing location, that is, $\mathcal{I}(h;\rho_{t+1},\mathbf{z}_{1:t+1}\mid\mathbf{x}_{1:t+1})$, as in~\cref{prob:homotopic_tracking}. More explicitly, we take the expectation of~\cref{eq:homotopic_KL} over the possible future partial $h$-signatures $\rho_{t+1}$ and beliefs obtained from observing ray crossing $\rho_t\rightarrow\rho_{t+1}$,
\begin{equation}\label{eq:homotopy:kl_div}
\mathcal{I}(h; \rho_{t+1}, \mathbf{z}_{1:t+1}\mid\mathbf{x}_{1:t+1}) = \mathbb{E}_{\rho_{t+1}}[D_{\mathrm{KL}}(b_{t},b_{t+1})].
\end{equation}

The expectation breaks down as follows,
\begin{multline}
    \mathbb{E}_{\rho_{t+1}}[D_{\mathrm{KL}}(b_{t},b_{t+1})] = \\ \sum_c\sum_{\rho_{t+1}}p(c)p(\rho_{t+1} \mid c)p(d\mid\mathbf{x}_t, c)D_\mathrm{KL}\big(b_{t},b_{t+1}\big),    
\end{multline}
where $p(\rho_{t+1} \mid c)$ is the probability of ray crossing $\rho_t \rightarrow \rho_{t+1}$ occurring at time $t$ given a GMM component $c$, $p(d\mid\mathbf{x}_t, c)$ is the probability of detecting the target at sensing location $\mathbf{x}_t$ given $c$ and $p(c)$ is the weight of the component.

The probability of a ray crossing $\rho_t \rightarrow \rho_{t+1}$ occurring at time $t$ given a GMM component $c$ is found by evaluating the cumulative distribution function of the component density on either side of the appropriate ray at times $t$ and $t+1$. The probability of detecting the target given the sensing location is modelled with a Bernoulli random field, where at each sensing location $\mathbf{x}_{t+1}$ the probability of detection is modelled by a Bernoulli random variable $d$.

To model the success parameter of the Bernoulli distribution we introduce a probabilistic detection model that decays exponentially as the distance between the robot and the target increases,
\begin{equation}
    p(d\mid \mathbf{x}_{t+1}, \mathbf{y}_{t+1}) = A \exp\left(-\frac{\lVert \mathbf{x}_{t+1} - \mathbf{y}_{t+1} \rVert^2}{2r^2}\right),
\end{equation}
where $r$ is the sensing radius of the robot and $A$ controls the peak of the model. However, since the target's future state $\mathbf{y}_{t+1}$ is unknown, marginalisation using the GMM is required to obtain a tractable result. The success parameter $\gamma^c(\mathbf{x}_{t+1})= p(d=\mathrm{True}\mid\mathbf{x}_{t+1},c)$ is then for each GMM component $c$,
\begin{equation}\label{eq:homotopy:component_wise_detect}
     \gamma^{c}(\mathbf{x}_{t+1}) = \beta\exp \left(-\frac{1}{2}{\Theta_{t+1}^{c}}^{\mathsf{T}}(r^2I + \Sigma_{t+1}^{c})^{-1}\Theta_{t+1}^{c}\right),
\end{equation}
 where $\boldsymbol{\mu}_{t+1}$, $\Sigma^{c}_{t+1}$ are the mean and covariance of the GMM component $c$ at $t+1$, $\Theta^{c}_{t+1} = \mathbf{x}_t-\boldsymbol{\mu}_{t+1}^{c}$, and $\beta= A\cdot(\det(\Sigma^{c}r^{-2} + I))^{-1/2}$ is a normalisation factor.

\subsection{Complexity analysis}\label{sec:homotopic_info:complexity}
Notably, the homotopic information gain is constructed from discrete sums and tractable cumulative distribution functions. At worst, the set $\mathcal{H}_{\rho_{t+1}}$ of full $h$-signatures compatible with a partial $h$-signature $\rho_{t+1}$ is the set of all possible $h$-signatures in the data. Then, the worst-case complexity of the homotopic information gain is $\mathcal{O}(MH^2)$, where $M$ is the number of GMM components and $H$ is the number of unique full $h$-signatures present in the training data. However, typically $\mathcal{H}_{\rho_{t+1}}$ is a subset of all possible $h$-signatures and complexity is lower.

\subsection{Theoretical analysis}
As seen in~\cref{eq:homotopy:kl_div}, the sensing robot does not take measurements of the target's partial $h$-signature $\rho_t$ directly, but rather of the target's position $\mathbf{y}_t$. Although one may consider sensor and filter designs that enable minimalist sensing of $\rho_t$ and corresponding expected information gains directly (see for example~\cite{lavalle_sensing_2010,siciliano_sensor_2009}), we wish to maintain measurements of position to help inform our overall estimate of the target's trajectory $\mathbf{Y}$. We merely plan over the homotopic belief space as a computationally sparse proxy for collecting information regarding $\mathbf{Y}$. A natural question is then how much information in $\mathbf{Y}$ is collected when planning sensing policies to maximise information in $h$.

One may expect that active target tracking using the homotopic information gain will prioritise high-level motion information only, ignoring the low-level information typically important for improving estimates of $\mathbf{Y}$. However, the following theorem establishes that when using the homotopic GMM in~\cref{eq:homotopy:homotopic_gmm} to model and predict trajectories, maximising homotopic information gain simultaneously maximises low-level information gain in the original~\cref{prob:active_perception}.
\begin{theorem}\label{theorem:info}
    Homotopic information gain is a lower bound for low-level, metric information gain. That is,
    \begin{equation}
        \mathcal{I}(h; \rho_t, \mathbf{z}_{1:t} \mid \mathbf{x}_{t}) \leq \mathcal{I}(\mathbf{Y};\rho_t, \mathbf{z}_{1:t}  \mid\mathbf{x}_t), 
    \end{equation}
    where $\mathcal{I}(\mathbf{Y};\rho_t,\mathbf{z}_{1:t})$ is the expected low-level information gained from measurements $\mathbf{z}_{1:t}$ and partial $h$-signature $\rho_t$ regarding the full trajectory $Y$.
\end{theorem}

This statement follows from recognising that the $h$-signature is a function of the full trajectory $\mathbf{Y}$. Then, an application of the data processing inequality~\cite{el_gamal_network_2011} gives the result. Intuitively, this result reflects the fact that no pre-processing of raw data can ever increase information gain under any circumstances. We note that the above bound applies only when using the homotopic GMM to model $\mathbf{Y}$, such that relevant topological information such as partial $h$-signature, obstacle numbers and locations are inherently present in the low-level model used to calculate metric information gain.

\section{Planning framework}

With an understanding of the construction and nature of homotopic information gain, we now propose a framework for solving~\cref{prob:homotopic_tracking}. 

\subsection{Active target tracking as an orienteering problem}
Due to the dynamic nature of the target, the information gain in~\cref{eq:homotopy:kl_div} is also dynamic. At a given sensing location, the information gain is only high for the period of time that the target is likely to be observed there. We therefore formulate the active target tracking problem as an orienteering problem with time windows (OPTW)~\cite{kantor_orienteering_1992}. This formulation introduces flexibility in the time frame in which the sensing robot can visit a sensing location, allowing for small uncertainty in the target's time-of-arrival at a sensing location to be handled.

A solution to the OPTW problem selects locations to visit within pre-defined time windows and an order to visit them in such that time constraints are satisfied and the expected reward is maximised. To define the locations, time windows and reward of the OPTW, we compute the expected information gain at discrete sensing locations placed on a uniform grid for all times from the current to the final $T=100$ in $1$s intervals. 

\begin{figure}[t]
    \centering
    \subfloat[Full heatcube.]{\includegraphics[width=0.48\columnwidth]{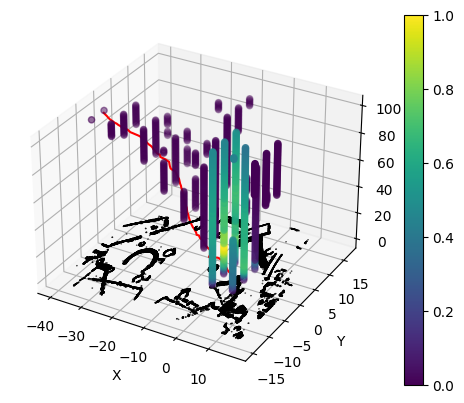}\label{fig:homotopy:full_kl}}\hfill
    \subfloat[Thresholded heatcube.]{\includegraphics[width=0.48\columnwidth]{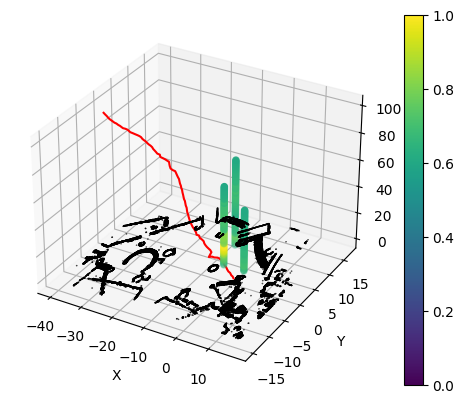}\label{fig:homotopy:blob_kl}}
    \caption{Heatcubes of homotopic information gain at different sensing locations and times ($z$-axis). Red line shows the test trajectory travelling upwards in time.}
    \label{fig:enter-label}
\end{figure}

A heatcube constructed with the homotopic information gain in~\cref{eq:homotopy:kl_div} for the ATC dataset is depicted in~\cref{fig:homotopy:full_kl}. The homotopic information gain is sparsely distributed in space and time -- it is highest around the first obstacle during the time window that the target is likely to move past this obstacle. We further threshold the heatcube to keep only the highest values. For example, in~\cref{fig:homotopy:blob_kl} the heatcube is thresholded at $0.7$ leaving only three sensing locations that persist along the time axis.
The location of these remaining features together with the time windows during which they exist define the OPTW to be solved at each planning step. Heatcubes and OPTW definitions for other information gain metrics can be constructed in the same manner. The time complexity of construction is $\mathcal{O}(T_p*S*X)$ where $X$ is the time complexity of the information gain calculation, $S$ is the number of discrete sensing locations, and $T_p$ is the remaining planning horizon.

We construct and explore a search tree over possible paths to find a solution to the OPTW defined by an information gain heatcube. To this end, we initialise the root node of the tree as the robot's initial state $\mathbf{x}_0$. Then, we simultaneously expand and explore the search tree for informative paths using a Monte Carlo Tree Search (MCTS) algorithm (\cref{alg:bg:mcts}). MCTS efficiently explores the search tree by prioritising expansion of nodes according to their promise. With each iteration of the algorithm, a node is \textit{selected} and \textit{expanded} by adding a child node to the tree. A child node is added to the tree only if the sensing robot can travel from the parent node to the child before its associated time window closes. At each node $v$, the sensing location $\mathbf{x}_t$ and the planned time of visit $t$ are stored. The expected information gain or reward of the newly added node is estimated via a simulation or \textit{roll-out} of a possible future sensing trajectory. The expected information gain $R$ is then \textit{back-propagated} and used in future iterations to calculate the promise of nodes for expansion.

\begin{algorithm}[t]
\textbf{Inputs:} Root node $v_0, \kappa>0$. 
\begin{algorithmic}[1]
    \While{termination condition not met}
        \State $v_{l} =$ \Call{selection}{$v_0$} \Comment{~\cref{eq:bg:uct}}
        \State $v^* =$ \Call{expansion}{$v_l$}
        \State $R =$ \Call{rollout}{$v^{*}$}
        \State \Call{backpropagation}{$R$}
    \EndWhile
    \State \Return maximum expected reward child of $v_0$
\end{algorithmic}
\caption{Monte Carlo Tree Search\label{alg:bg:mcts}}
\end{algorithm}

The notion of promise is captured by the upper confidence bound for trees (UCT) policy~\cite{Kocsis_bandit_2006}. That is, a child node $v^{\prime}$ of parent node $v$ is selected for expansion if it maximises the UCT,
\begin{equation}\label{eq:bg:uct}
    \mathrm{UCT}(v^{\prime}) = \underbrace{\frac{R(v^{\prime})}{N(v^\prime)}}_{\text{exploitation}} + \kappa \underbrace{\sqrt{\frac{2\ln N(v)}{N(v^\prime)}}}_{\text{exploration}},
\end{equation}
where $N(v)$ is the number of times the parent node has been visited, $R(v^\prime)$ is the cumulative reward and $\kappa > 0$ is a constant that dictates how much exploration is allowed. In the context of this active tracking problem, the cumulative reward is the sum of expected information gain over the planning horizon, precomputed in the construction of the heatcube. The sum of information gain along a trajectory calculated with a fixed belief overestimates the true information gain. However, since all roll-outs suffer overestimation and only the next sensing location is taken from the MCTS solution before replanning, the final information gained approximates the objective function well. Nevertheless, expansion is guided by balancing selection of nodes with high expected information gain and nodes that have not been visited often or at all (a node with $N(v^\prime)=0$ has infinite UCT).

After a user-defined number of roll-outs to the end of the remaining planning horizon, the estimated maximally informative next action is taken and a measurement is acquired. Then, the belief is updated and re-planning over an updated heatcube to solve a new OPTW is performed until the target trajectory terminates.

\subsection{Online belief updates}\label{sec:gmm}

Here we specify how the homotopic GMM is updated with online measurements. By introducing a sensing robot we introduce an added uncertainty: whether or not the target will be detected at $\mathbf{x}_{t}$ and a measurement will be received. To adjust the belief updates accordingly, we scale the weights by the probability of detection in~\cref{eq:homotopy:component_wise_detect}. That is, when the target is detected at $\mathbf{x}_{t}$, weights are updated as
\begin{equation}\label{eq:online_update_detect}
    \hat{w}^{h'} \propto \gamma^{h'}(\mathbf{x}_{t})w^{h'}P(h\mid\rho_{t})\mathcal{N}(\boldsymbol{\mu}^{h'},\Sigma^{h'} + \sigma_Z^2I),
\end{equation}
where the hat $\hat{(\cdot)}$ indicates a posterior quantity, $p(h'\mid \rho_{t})$ is given by the homotopic belief and $\sigma_Z$ is measurement noise variance,
and when a target is not detected at a sensing location $\mathbf{x}_{t}$,
\begin{equation}\label{eq:online_update_misdetect}
    \hat{w}^{h'} \propto (1-\gamma^{h'}(\mathbf{x}_{t}))w^{h'}P(h'\mid\rho_{t})\mathcal{N}(\boldsymbol{\mu}^{h'},\Sigma^{h'}).
\end{equation}

\begin{figure*}[t]
    \centering
    \begin{tabular}{ccc}
         \subfloat[A simulated dataset with three obstacles of varying sizes.]{\includegraphics[width=0.3\linewidth]{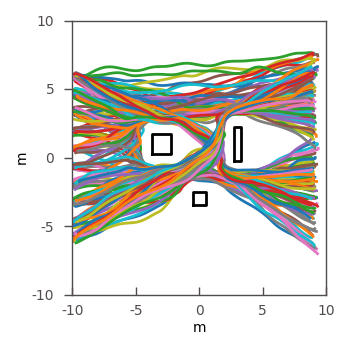}\label{fig:sim_data_full}}  &
         \subfloat[Full ATC dataset. The 5 obstacles present in the environment are filled in black~\cite{atc_dataset}.]{\includegraphics[width=0.3\linewidth]{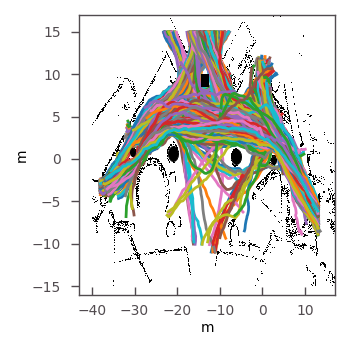}\label{fig:atc_data_full}} &
         \subfloat[Full TH\"{O}R dataset with augmented trajectories~\cite{thorDataset2019}.]{\includegraphics[width=0.3\linewidth]{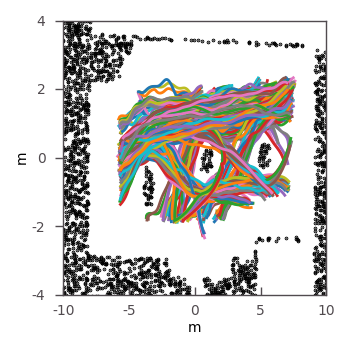}\label{fig:thor_data_full}}\\
         \subfloat[Homotopy classes present in the three obstacle simulated data. Two samples from each class are shown to illustrate multi-modal behaviour within homotopy classes. Colours indicate homotopy class.]{\includegraphics[width=0.3\linewidth]{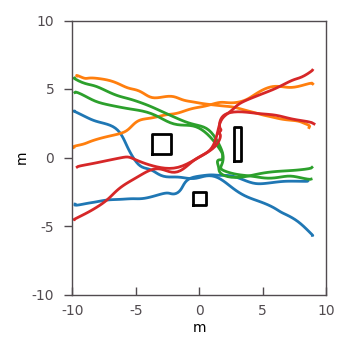}\label{fig:sim_data_hsig}}&
         \subfloat[Homotopy classes present in the ATC dataset. Two samples from each class are shown to illustrate multi-modal behaviour within homotopy classes. Colours indicate homotopy class.]{\includegraphics[width=0.3\linewidth]{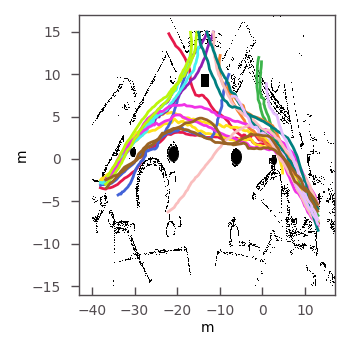}\label{fig:atc_data_hsig}}&
         \subfloat[Homotopy classes present in the TH\"{O}R dataset. Two samples from each class are shown to illustrate multi-modal behaviour within homotopy classes. Colours indicate homotopy class.]{\includegraphics[width=0.3\linewidth]{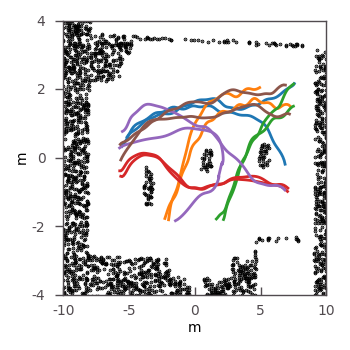}\label{fig:thor_data_hsig}}
    \end{tabular}
    \caption{Left: one of the environments used to simulate pedestrian trajectories, here with three obstacles. Middle: The real pedestrian ATC mall dataset with five obstacles present. Right: the TH\"{O}R dataset with both real and augmented pedestrian trajectories with three obstacles present. }
    \label{fig:datasets}
\end{figure*}

\section{Evaluation of homotopic information gain for active target tracking}
We compare our proposed approach to a conventional information gathering approach, which aims to maximise the \textit{metric} or low-level information gain provided by measurements by maximising conditional entropy of the posterior GMM.

We implement a naive measure of low-level information gain to avoid the overhead associated with the aforementioned pairwise distance estimates of entropy~\cite{kolchinsky_estimating_2017}. That is, we implement the weighted sum of conditional entropies of each Gaussian in the measurement GMM scaled by the probability of detection at $\mathbf{x}_t$. The metric information gain used is therefore
\begin{equation}\label{eq:homotopy:naive_information_gain}
     \mathcal{I}(\mathbf{y}_t; \mathbf{z}_t) = p(d\mid \mathbf{x}_t) \cdot \sum_{h} \hat{w}_t^{h}\log \det\left(\hat{\Sigma}_t^{h} + R^{h}(\mathbf{x}_t)\right), 
\end{equation}
where $R^{h}(\mathbf{x}_t)$ is the measurement noise covariance, modelled as a function of the distance between $\mathbf{x}_t$ and the mean of component $h$. This simplification reduces the computational complexity of evaluating the metric information gain at a single location from $\mathcal{O}(M^2N^3)$ to $\mathcal{O}(MN^3)$, where $M$ is the number of GMM components and $N=2T$ is the size of $\Sigma_t$. Since many locations must be considered for many time steps during planning, this is a significant reduction, particularly when the GMM has many components.

For each method three metrics are evaluated for comparison. First we consider the displacement error at each timestep, DE$(t)$. This measures the average distance between the ground truth trajectory and the highest-weighted mean of the final posterior GMM after the active target tracking mission is complete.
This is defined as
\begin{equation}\label{eq:ADE}
    \text{DE}(t) = || \mathbf{x}_{t} - \boldsymbol{\hat{\mu}}_{t}^{ c^{*}}||,
\end{equation}
where $c^{*}$ denotes the index of the component with the highest weight.  

To perform further evaluation we introduce a homotopic GMM conditioned on the \textit{full} test trajectory, referred to as the ground truth GMM. Then, we evaluate the \textit{Kullback-Leibler divergence} (KLD) between the weights of the ground truth GMM, $\hat{w}_{T}^{(c,h)}$, and the weights of the partial posterior GMM $\hat{w}_{\mathrm{obs}}^{(c,h)}$ after each measurement is taken. This quantifies the high-level information gain of each approach on a per-measurement basis.
\begin{equation}\label{eq:KLD}
    \text{KLD} = \sum_{c, h}  \hat{w}_{T}^{ (c, h) } (\log \hat{w}_{T}^{ (c, h) } - \log \hat{w}_{\mathrm{obs}}^{ (c, h) }).
\end{equation}

Finally, we study the low-level information gain per measurement by evaluating the mutual information between the ground truth GMM and the partial posterior GMM post measurement. To do so, we evaluate the variational upper bound on the mutual information between two GMMs presented in~\cite{hershey_approximating_2007}. For full posterior GMM $f$ with weights and Gaussian components $(w_a,f_a)$, and partial posterior GMM $g$ with $(w_b, g_b)$ this is given by
\begin{equation}
    D_{\mathrm{var}}(f,g) = \sum_a w_a \log \frac{\sum_{a'}w_{a'}\exp(-D_{\mathrm{KL}}(f_a,f_{a'}))}{\sum_{b}w_b\exp(-D_{\mathrm{KL}}(f_a,g_b))}.
\end{equation}

To evaluate the efficiency of our homotopic framework compared to the metric approach we study total runtime. Computation of the heatcubes used for the definition of the OPTW and reward function calculation is not constant at each re-planning step, as the planning horizon shortens as the trajectory progresses. Furthermore, MCTS compute time is not constant due to the dynamic search space dimension at each re-planning step. For fair comparison we therefore report the time taken for the entire tracking experiment to complete. All tests are run on an Intel 3.2GHz processor.

\begin{table}[t]
    \centering
    \begin{tabular}{c|ccccc|c|c}
         & \multicolumn{7}{c}{Dataset} \\
         & \multicolumn{5}{c|}{Simulated} & TH\"{O}R & ATC\\
         \hline \rule{0pt}{2.5ex}
         $O$ & 3 & 4 & 5 & 6 & 7 & 3 & 5 \\
         $H$ & 4 & 6 & 7 & 11 & 12 & 6 & 13
    \end{tabular}
    \caption{The number of obstacles $O$ and homotopy classes $H$ present in each dataset used for evaluation.}
    \label{tab:hsigs}
\end{table}

\subsection{Datasets}
We evaluate our methodology on both simulated and real pedestrian datasets, detailed below. The number of distinct homotopy classes $H$ and obstacles $O$ present in each dataset is tabulated in~\cref{tab:hsigs}. In all datasets, pedestrians begin and end on boundary points. All trajectories were interpolated over $100$ discrete timesteps to fix the dimension of the GMM components and enable prediction of the full trajectory. To begin, we exclude the homotopy class passing under all obstacles from all datasets. The partial $h$-signature for this class remains empty for the whole trajectory as it does not pass any rays emanating from obstacles. This therefore results in no change in the homotopic belief output by the VOMP. Thus, comparing two identical beliefs when evaluating~\cref{eq:homotopy:kl_div} results in zero homotopic information gain, limiting the use of our method. To quantify the impact of the empty signature set on the performance of our framework we evaluate again on all datasets with the empty signature class re-introduced in~\cref{sec:empty_sig_ablate}.

\subsubsection{Simulated datasets} To test the scalability of our method, a series of simulated datasets were created using the crowd simulation engine in~\cite{karamouzas17} with an increasing number of obstacles in the environment, ranging from $3$ up to $7$. In all simulated environments, the pedestrians enter the environment on the left boundary and exit on the right boundary, travelling from left to right. An example of one simulated environment is shown in~\cref{fig:sim_data_full}. The four distinct homotopy classes remaining are shown in~\cref{fig:sim_data_hsig}). For each dataset, 48 training trajectories per homotopy class were randomly selected to form a training set, and 12 from each were selected to build a test set. The number of components per homotopy class in the homotopic GMM was $N_C = 1$.

\subsubsection{ATC dataset} We evaluate on real world trajectories from the ATC shopping mall dataset~\cite{atc_dataset} to test our approach on more accurate human walking behaviour. From the ATC dataset we extracted a subdomain of the full shopping mall environment with 5 obstacles within. Then, trajectories that were fully contained within the chosen subdomain were selected from the dataset. As seen in~\cref{fig:atc_data_full}, the real-world data has more varied low-level motions than the simulated data. Here pedestrians travel in all directions rather than strictly from left to right. The 13 most frequent homotopy classes in the data are depicted in~\cref{fig:atc_data_hsig}, illustrating the diversity of high-level motions present in the dataset. We randomly selected 1300 train and 260 test trajectories from this set. The number of components per homotopy class in the homotopic GMM was $N_C=3$, resulting in a total of $39$ components.

\begin{figure*}[t]
    \centering
    \subfloat[Frequency count of total measurements taken by each method per target tracking experiment.]{\includegraphics[width=0.32\textwidth]{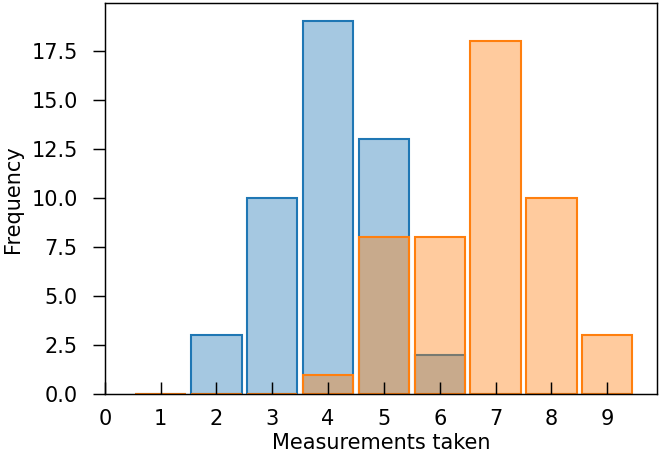}\label{fig:results:sims:histo}}\hfill
    \subfloat[Displacement error. Solid line is the average displacement error. Shaded region shows the  interquartile range.]{\includegraphics[width=0.32\textwidth]{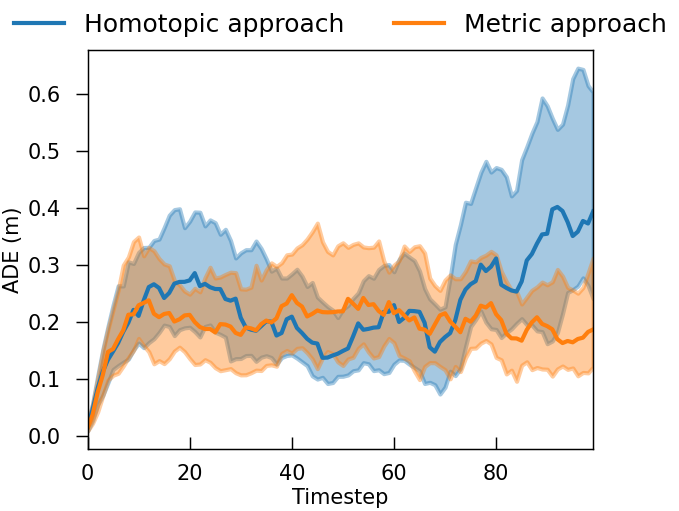}\label{fig:results:sims:ade}}\hfill
    \subfloat[High-level (top) and low-level (bottom) information gained per measurement.]{\includegraphics[width=0.32\textwidth]{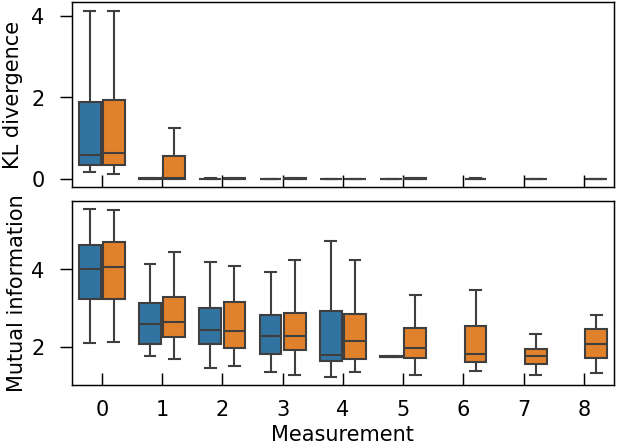}\label{fig:results:sims:info}}
    \caption{Results for active target tracking on the simulated pedestrian dataset with three obstacles.}\label{fig:results:sims}
\end{figure*}

\begin{figure*}[t]
    \centering
    \subfloat[Number of measurements taken per tracking experiment for each approach.]{\includegraphics[width=0.32\textwidth]{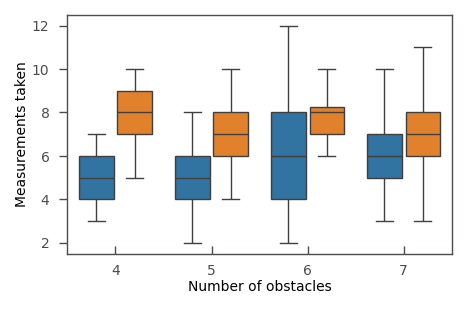}\label{fig:results:sims:ablation:measurement}}\hfill
    \subfloat[The displacement error averaged over the full trajectory length for each simulated dataset.]{\includegraphics[width=0.32\textwidth]{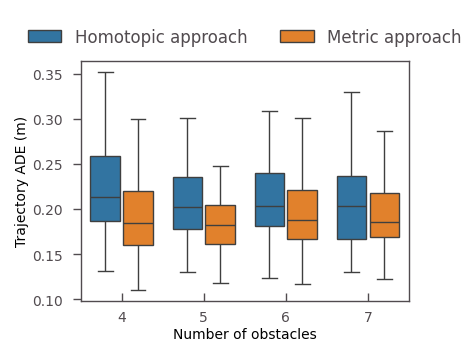}\label{fig:results:sims:ablation:ade}}\hfill
    \subfloat[Total tracking experiment length for increasing number of obstacles in the simulated datasets.]{\includegraphics[width=0.32\textwidth]{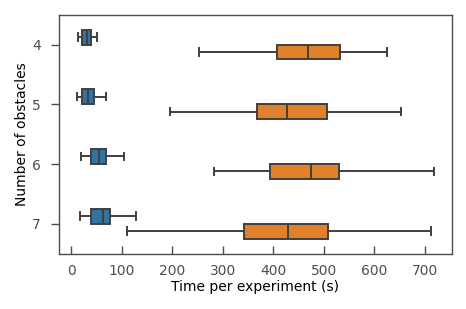}\label{fig:results:sims:ablation:compute}}
    \caption{Results for active target tracking on the simulated pedestrian datasets increasing number of obstacles.}\label{fig:results:sims:ablation}
\end{figure*}

\begin{figure*}[t]
    \centering
    \subfloat[Frequency count of total measurements taken by each method per target tracking experiment.]{\includegraphics[width=0.32\textwidth]{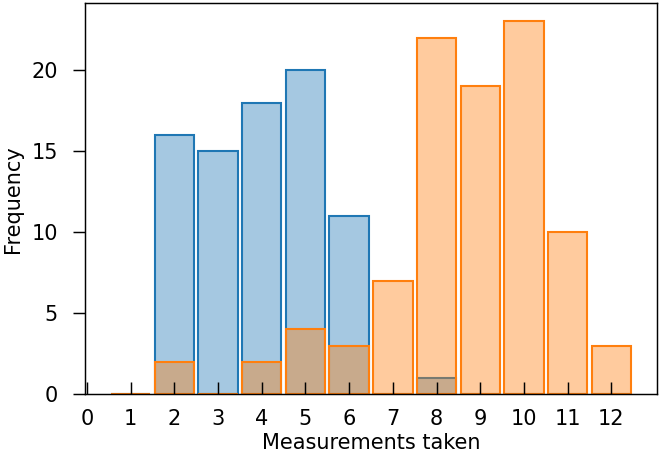}\label{fig:results:atc:histo}}\hfill
    \subfloat[Displacement error. Solid line is the average displacement error. Shaded region shows the  interquartile range.]{\includegraphics[width=0.32\textwidth]{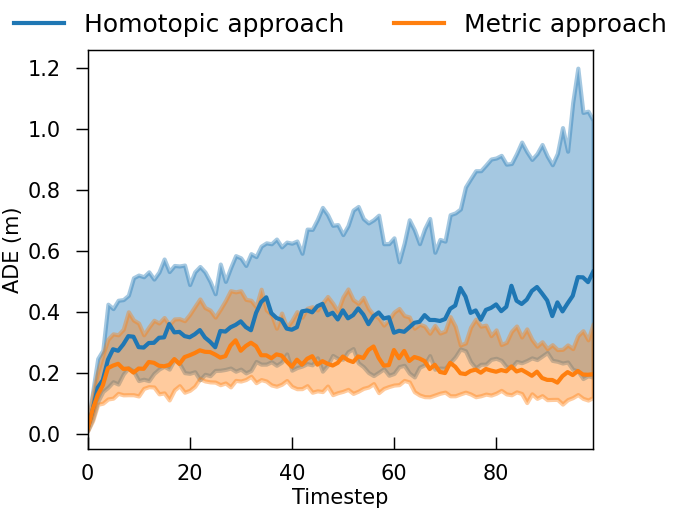}\label{fig:results:atc:ade}}\hfill
    \subfloat[High-level (top) and low-level (bottom) information gained per measurement.]{\includegraphics[width=0.32\textwidth]{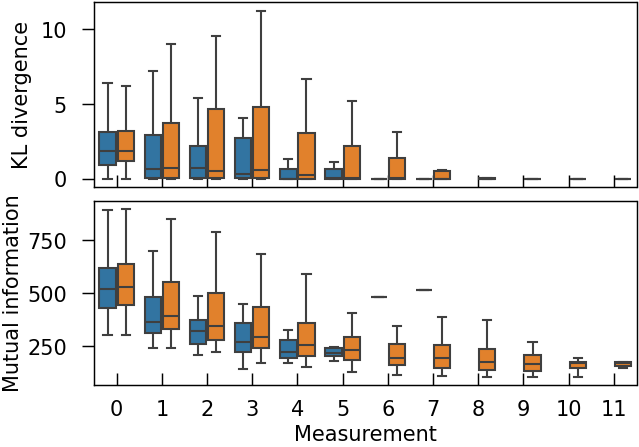}\label{fig:results:atc:info}}
    \caption{Results for active target tracking on the ATC mall pedestrian dataset with five obstacles.}\label{fig:results:atc}
\end{figure*}

\begin{figure*}[t]
    \centering
    \subfloat[Frequency count of total measurements taken by each method per target tracking experiment.]{\includegraphics[width=0.32\textwidth]{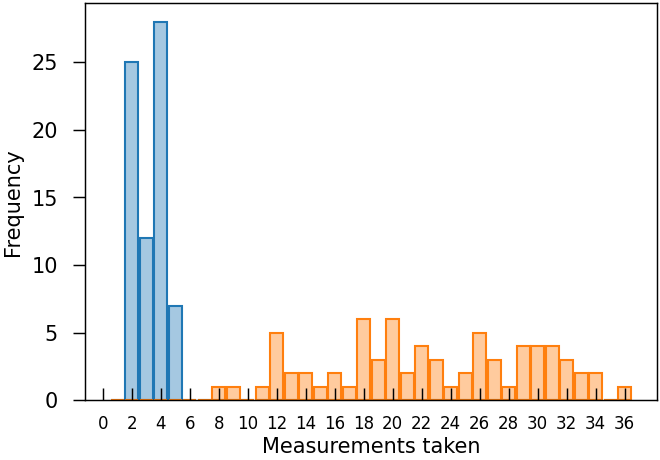}\label{fig:results:thor:histo}}\hfill
    \subfloat[Displacement error. Solid line is the average displacement error. Shaded region shows the  interquartile range.]{\includegraphics[width=0.32\textwidth]{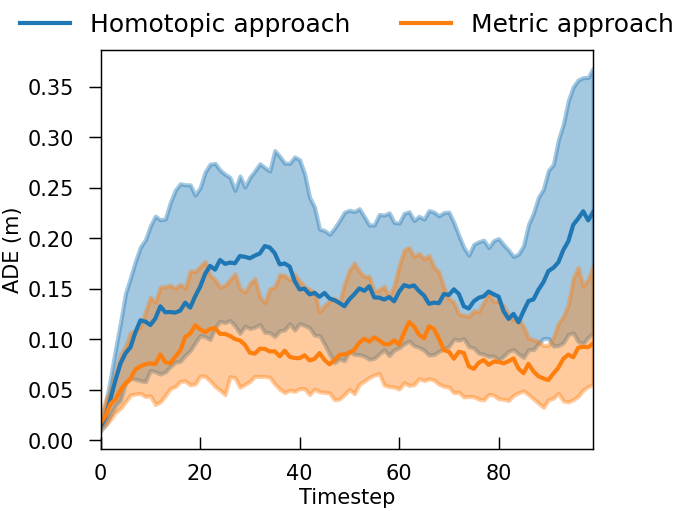}\label{fig:results:thor:ade}}\hfill
    \subfloat[High-level (top) and low-level (bottom) information gained per measurement.]{\includegraphics[width=0.32\textwidth]{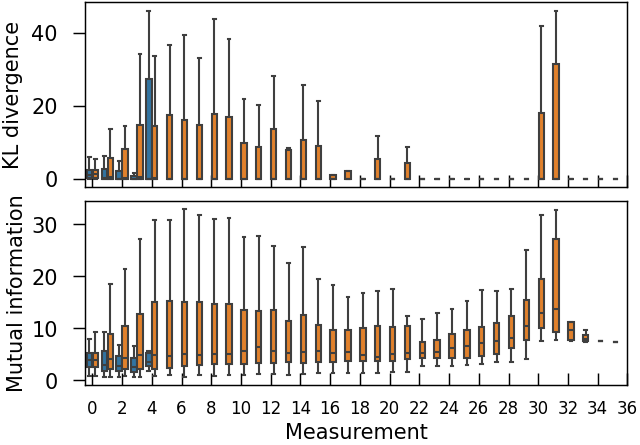}\label{fig:results:thor:info}}
    \caption{Results for active target tracking on the augmented TH\"{O}R pedestrian dataset with three obstacles.}\label{fig:results:thor}
\end{figure*}

\begin{figure}[t]
    \centering
    \includegraphics[width=\columnwidth]{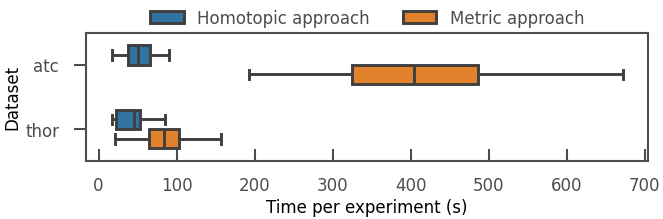}
    \caption{Total tracking experiment length for the real datasets.}
    \label{fig:results:real_data_compute}
\end{figure}

\subsubsection{TH\"{O}R dataset} A dataset of real pedestrians navigating in all directions to and from fixed goals on the boundary of the environment is publicly available in~\cite{thorDataset2019}. We take the dataset with $3$ obstacles for our evaluation. To increase the number of trajectories in the dataset and facilitate a test-train split, trajectories that navigate between $3$ or more goals are split into sub-trajectories that only navigate between two goal locations. After this filtering, the number of trajectories in each distinct homotopy class is still small. We therefore augment the dataset with statistically similar trajectories by fitting a Gaussian process to original trajectories and sampling the resulting posterior. The resulting $288$ trajectories in the training set are shown in~\cref{fig:thor_data_full}, with corresponding $6$ homotopy classes in~\cref{fig:thor_data_hsig}. $72$ trajectories were saved for testing. This dataset contained the most multi-modal behaviour within homotopy classes of all datasets. More than topological complexity, this dataset has significant low-level complexity compared to the other datasets. Trajectories in this dataset also have high-frequency low-level variations, as seen in blue and orange sample trajectories in~\cref{fig:thor_data_hsig}. We therefore trained the homotopic GMM with $N_C = 4$ components per class.

\subsection{Results}

\subsubsection{Simulated datasets} Evaluation of the homotopic active target tracking approach against the metric on the $3$ obstacle simulated dataset is shown in~\cref{fig:results:sims}. In~\cref{fig:results:sims:ade}, the average displacement error (ADE) over all test trajectories is plotted with the interquartile range. Viewed together with~\cref{fig:results:sims:histo} we see our homotopic approach achieves competitive accuracy in terms of ADE with far fewer measurements than the metric approach -- taking a median of just 4 measurements compared to 7. The high-level information gained per measurement shown in~\cref{fig:results:sims:info} reflects that for low-complexity data, there is little uncertainty in the homotopy class of a trajectory. However, the low-level mutual information between the ground truth GMM and the posterior GMM for our homotopic approach reaches a lower value after only 4 measurements than the metric approach does in 9. This corroborates~\cref{theorem:info}, showing that despite being designed to maximise high-level information, the homotopic planning approach collects highly informative measurements for low-level prediction as well. 

To study the scalability of our approach with increasing number of obstacles, a comparison of the runtime per test trajectory for each simulated dataset is shown in~\cref{fig:results:sims:ablation:compute}. Here we see for an increasing number of obstacles and thus homotopy classes, the computation of our proposed approach does indeed increase, corroborating the analysis in~\cref{sec:homotopic_info:complexity}. However, the runtime for the proposed approach even for $H=12$ is still far less than the metric approach. \cref{fig:results:sims:ablation:measurement} shows for each dataset the homotopic approach again takes far fewer measurements than the metric while maintaining impressive ADE in~\cref{fig:results:sims:ablation:ade}, where the reported ADE is as in~\cref{eq:ADE} but averaged over all time steps.

For environments up to $6$ obstacles, both methods were $100$\% successful at completing a tracking experiment. We consider an experiment successful if the sensing robot measures the agent at at least one location after the initial measurement at $t=0$. For the $6$ obstacle environment, the homotopic approach is $99$\% successful while the metric is $100$\% successful. For the $7$ obstacle environment, the success rates are $91$\% for the homotopic and $100$\% for the metric approach. This drop in success rate as the number of obstacles increases is due to the greater variance in trajectory lengths for datasets with more complex topological trajectories -- it takes longer for a pedestrian to weave through many obstacles than it does to pass above them. Since we interpolate all trajectories such that they have a fixed number of timesteps $T=100$, this introduces a large variance in the average velocity of trajectories in the dataset. As a result, there is great time uncertainty introduced to the OPTW problem that is not robustly handled in our framework. This disproportionately impacts the homotopic approach as the belief space is sparser than the metric, resulting in only a few sensing locations being considered over the uncertain time window. Meanwhile, the metric approach plans to take measurements densely in both space and time, essentially `following' the agent rather than moving ahead and waiting at a future obstacle.

\begin{figure*}[t]
    \centering
    \begin{tabular}{cc}
        \subfloat[Ours $t=1$]{\includegraphics[width=0.19\textwidth]{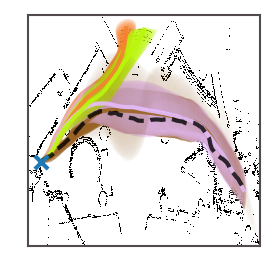}\label{fig:tracking:hig_1}} 
        \subfloat[Ours $t=35$]{\includegraphics[width=0.19\textwidth]{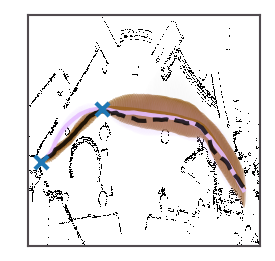}\label{fig:tracking:hig_2}} 
        \subfloat[Ours $t=65$]{\includegraphics[width=0.19\textwidth]{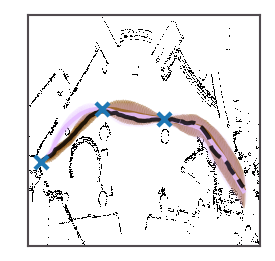}\label{fig:tracking:hig_3}} 
        \subfloat[Ours $t=80$]{\includegraphics[width=0.19\textwidth]{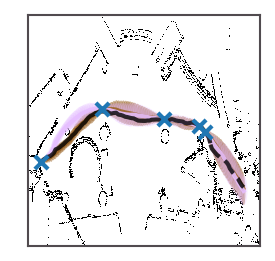}\label{fig:tracking:hig_4}} 
        \subfloat[Ours $t=100$]{\includegraphics[width=0.19\textwidth]{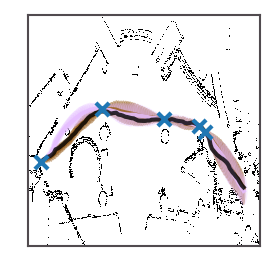}\label{fig:tracking:hig_5}} \\
        \subfloat[Metric $t=1$]{\includegraphics[width=0.19\textwidth]{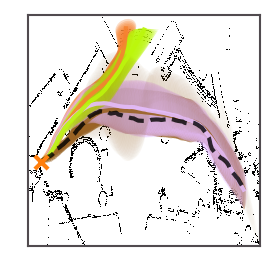}\label{fig:tracking:mig_1}} 
        \subfloat[Metric $t=35$]{\includegraphics[width=0.19\textwidth]{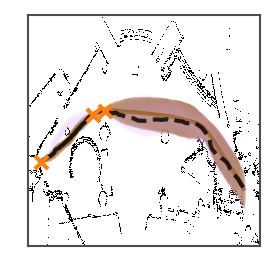}\label{fig:tracking:mig_2}} 
        \subfloat[Metric $t=65$]{\includegraphics[width=0.19\textwidth]{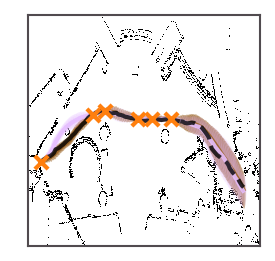}\label{fig:tracking:mig_3}} 
        \subfloat[Metric $t=80$]{\includegraphics[width=0.19\textwidth]{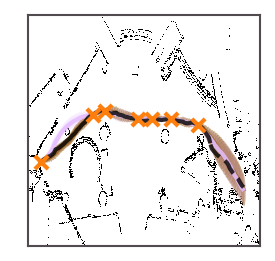}\label{fig:tracking:mig_4}} 
        \subfloat[Metric $t=100$]{\includegraphics[width=0.19\textwidth]{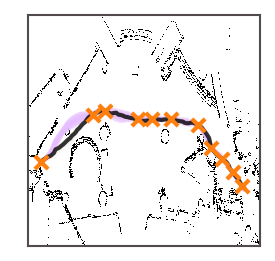}\label{fig:tracking:mig_5}}
    \end{tabular}
    \caption{Snapshots of predictions output by a homotopic GMM during an active target tracking experiment using homotopic information gain~\protect\subref{fig:tracking:hig_1}~-~\protect\subref{fig:tracking:hig_5} and metric information gain~\protect\subref{fig:tracking:mig_1}~-~\protect\subref{fig:tracking:mig_5}. Black solid path is the elapsed ground truth test trajectory at the given time. Dotted path indicates future ground truth trajectory remaining. Coloured crosses denote all measurements taken by the sensing robot up to time $t$, where colours serve to differentiate between approaches (orange for the metric approach, blue for the homotopic approach). Shaded regions show the variance of the mean shown in solid colour, where colours differentiate GMM components. Transparency is proportional to component weight.}
    \label{fig:tracking}
\end{figure*}

\subsubsection{ATC dataset} Active target tracking results for the real pedestrian data is shown in~\cref{fig:results:atc}. Again, utilising homotopic information gain in the active target tracking problem achieves competitive ADE with distinctly fewer measurements, as evidenced in~\cref{fig:results:atc:histo,fig:results:atc:ade}. Remarkably, the homotopic approach ADE in~\cref{fig:results:atc:ade} is achieved with a median of 4 measurements compared to 9 for the metric approach. With this more complex dataset, each measurement provides more high-level information until the homotopy class is completely determined, at which point the KLD converges to zero in~\cref{fig:results:atc:info}. Meanwhile, measurements collected by the metric approach do not necessarily elucidate the homotopy class of the trajectory giving belief updates that weigh incorrect motion hypotheses highly, resulting in higher KLD. By studying the mutual information measure in~\cref{fig:results:atc:info} together with~\cref{fig:results:atc:ade} we again see that the homotopic approach is highly effective at collecting informative measurements for low-level prediction. The persistence of this result in this challenging real dataset is strong empirical corroboration of the bound in~\cref{theorem:info}.

An example test trajectory is shown in~\cref{fig:tracking} to provide further intuition for how each information gain measure impacts active target tracking. In~\crefrange{fig:tracking:mig_1}{fig:tracking:mig_5}, we see that the metric approach takes a dense set of measurements distributed evenly along the length of the trajectory, attempting to collect as much positional information as possible. Meanwhile, the homotopic approach shown in~\crefrange{fig:tracking:hig_1}{fig:tracking:hig_5} takes sparse measurements localised around obstacles in the environment. These behaviours generalise beyond this specific trajectory; in~\cref{fig:spatial_heatmap} a spatial heatmap of sensing locations is depicted reflecting the same. Our proposed homotopic approach thus visits a much sparser set of locations concentrated around obstacles, a direct result of planning with an information gain measure that is sparsely distributed in space. In~\cref{fig:results:real_data_compute} we see the influence this has on the time required to complete a successful mission. Runtime is vastly lower for our approach compared to the metric approach on this real dataset.

The success rate of the homotopic approach for this dataset is $81$\% while for the metric approach it is $95$\%. In this dataset there is even greater time uncertainty and we see an impact on the success rate of the metric approach as well, supporting our intuition that this drawback is likely due to the shared planning framework more than the gain measure used. 

\subsubsection{TH\"{O}R dataset} Evaluation on the TH\"{O}R dataset is shown in~\cref{fig:results:thor}. For this dataset we see the largest discrepancy between methods for the number of measurements taken in~\cref{fig:results:thor:histo}.  Due to the higher low-level trajectory complexity in this dataset, the metric information approach takes substantially more measurements in this dataset (a median of 22). On the other hand, as homotopic information gain is localised around obstacles the median number of measurements by our proposed approach is only 3. Still, the ADE in~\cref{fig:results:thor:ade} for our method is competitive. We observe that both methods struggle to collect low-level information in~\cref{fig:results:thor:info} with the homotopic approach still outperforming the metric. Our approach again reduces high-level uncertainty at a faster rate than the metric. For experiments with 5 measurements the final KLD reported in~\cref{fig:results:thor:info} has a median of 0 but high variance, suggesting for some data points the final KLD is high. Investigation of these cases reveal a trend. The homotopic approach occasionally predicts $h=(1,2)$ class trajectories to be $h=(1,2,3)$ when the trajectory is not measured passing below the third obstacle. Due to the high number of measurements taken, the posterior GMM obtained via the metric approach miscategorises the homotopy class of the final full trajectory less often. The success rate for both methods on the TH\"{O}R dataset is 100\%.

Comparison of runtime for each method is shown in~\cref{fig:results:real_data_compute}. While our approach is still faster, here we see more competitive runtime for the metric approach than in other datasets. This is again due to the large low-level variance in the trajectories. Since the metric approach prioritises reducing this uncertainty, the resulting OPTW has a small search space concentrated in a tight region near the target's last observed location. With a reduced search space MCTS finds the next best sensing location faster, reducing overall experiment length.

\subsubsection{Empty signature ablation}\label{sec:empty_sig_ablate}
The choice of homotopy invariant described in~\cref{sec:bg:invariants} limits our approach to work only on paths that cross obstacle rays. Paths that do not cross obstacle rays have $h$-signature $h=()$. For this class, the partial $h$-signature does not change throughout the tracking experiment and the homotopic belief output by the VOMP will be static for all time. The homotopic information gain therefore evaluates to zero, incorrectly indicating a lack of homotopic information. However, the robot's sensing trajectory is guided by maximisation of \emph{expected} homotopic information gain, which is often non-zero due to uncertainty in measurements and the predicted next partial $h$-signature $\rho_{t+1}$ given sensing location $\mathbf{x}_{t+1}$. Then, to investigate the impact of this theoretical inability to capture information gain where $h=()$ we study the ADE, number of measurements taken and success rate for all datasets where the empty $h$-signature class has been reintroduced.

\begin{figure}[t]
    \centering
    \includegraphics[width=\linewidth]{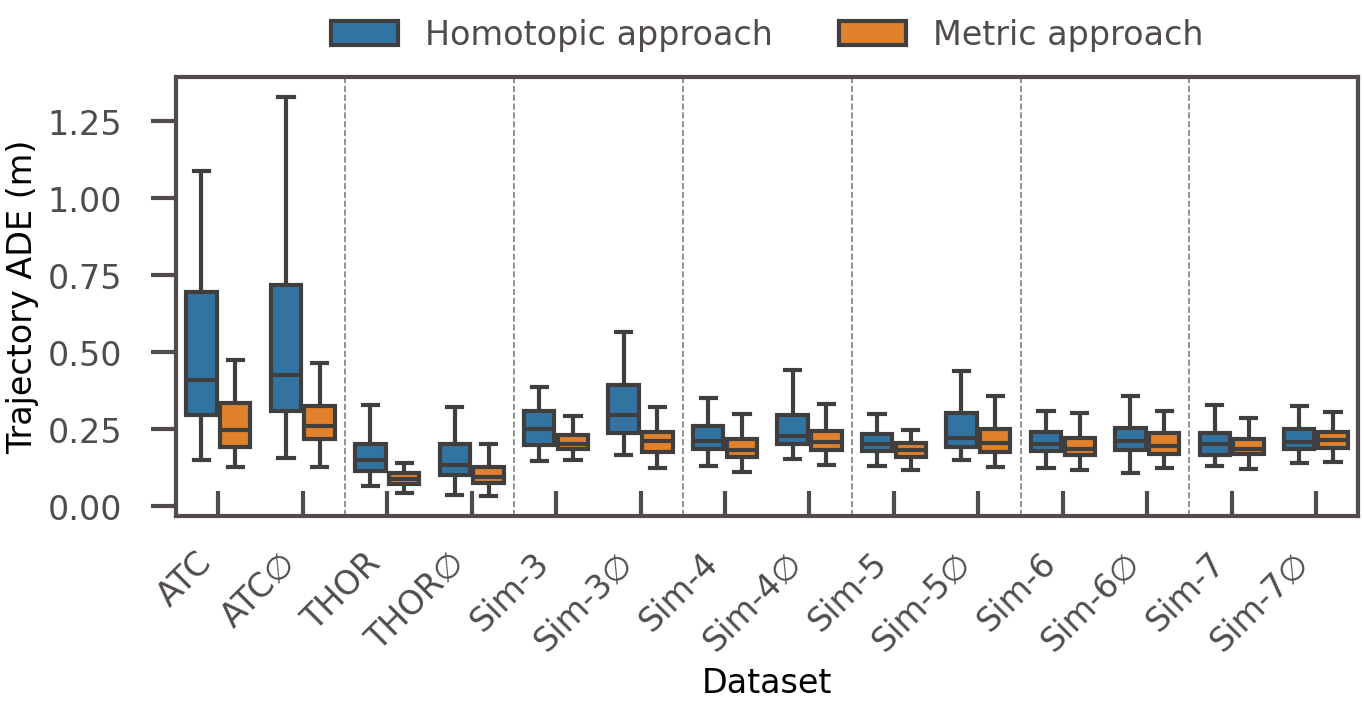}
    \caption{Distributions of displacement errors averaged over the full trajectory length for each dataset with (labelled with $\varnothing$) and without the empty $h$-signature.}
    \label{fig:empty_sig_ADE}
\end{figure}

\begin{figure}[t]
    \centering
    \includegraphics[width=\linewidth]{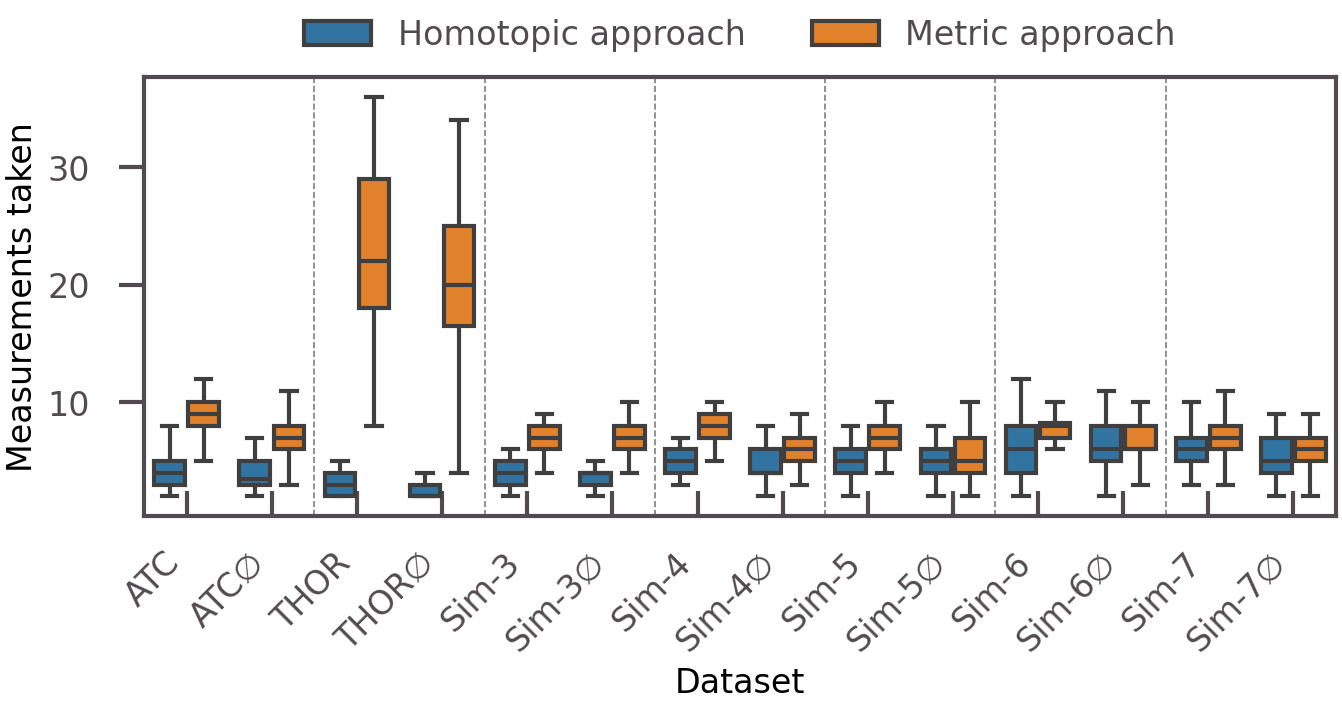}
    \caption{Distributions of the number of measurements taken per tracking experiment for each dataset with (labelled with $\varnothing$) and without the empty $h$-signature.}
    \label{fig:empty_sig_num_msmts}
\end{figure}

Results for all datasets with and without the empty $h$-signature class are shown in~\cref{fig:empty_sig_ADE}~and~\cref{fig:empty_sig_num_msmts}. Inclusion of this class does not change trends observed in ADE or the number of measurements for either approach significantly. The most significant impact of the $h=()$ class is seen when investigating the success rates of each approach, given in~\cref{tab:empty_success_rate}. Interestingly, the success rate for the ATC dataset increases slightly when $h=()$ is added. However, the increased ADE variance and vastly reduced number of measurements compared to the dataset without $h=()$ indicates that the increase in `successful' tracking experiments is due to the robot taking $2$ measurements rather than $1$ before failing to track the target. This results in miscategorisation of the final homotopy class and a much larger ADE. Nevertheless, in general success rate decreases when the $h=()$ class is added, becoming more significant as the complexity of the dataset increases. This is due to the robot being unable to track the target after moving to an irrecoverable sensing position while planning under an incorrect homotopic belief. 

These results support our intuition that the $h=()$ class presents a challenge to our approach. Still, here we see our approach is able to handle this class with only small impact to success rate -- the sparse number of measurements taken and competitive final ADE echoes the findings from the datasets where $h=()$ is excluded. This is due to our treatment of the partial $h$-signature as a random variable together with maximisation of the \textit{expected} homotopic information gain, resulting in a non-zero planning heuristic even for empty partial $h$-signatures. Regardless, observation of the target below an obstacle should explicitly provide homotopic information to the robot and be utilised in planning. We therefore believe a more expressive choice of homotopy invariant can further improve the performance of our algorithm, particularly on the $h=()$ set. For example, the homotopy invariant presented in~\cite{sakcak_convex_hsig} identifies homotopy class by considering the convex regions through which a path moves through rather ray crossings, and may provide non-empty labels for all trajectories.

\begin{table}[t]
\centering
\setlength{\tabcolsep}{4.5pt}  
\renewcommand{\arraystretch}{1.15}
\begin{tabular}{l|ccccccc}
\multicolumn{8}{c}{\textbf{Datasets without $h=()$}} \\
\hline
& Sim-3 & Sim-4 & Sim-5 & Sim-6 & Sim-7 & ATC & TH\"{O}R \\
\hline
Ours   & 100 & 100 & 100 & 99 & 91 & 81 & 100 \\
Metric & 100 & 100 & 100 & 99 & 98 & 95 & 100 \\
\multicolumn{8}{c}{}\\
\multicolumn{8}{c}{\textbf{Datasets with $h=()$}} \\
\hline
& Sim-3$\varnothing$ & Sim-4$\varnothing$ & Sim-5$\varnothing$ & Sim-6$\varnothing$ & Sim-7$\varnothing$ & ATC$\varnothing$ & TH\"{O}R$\varnothing$ \\
\hline
Ours   & 96.67 & 100 & 98.9 & 90 & 91 & 85 & 81 \\
Metric & 100 & 100 & 100 & 99 & 98 & 94 & 99 \\
\end{tabular}
\caption{Success rates (\%) for our homotopic approach versus the metric approach before and after the empty $h$-signature is introduced.}
\label{tab:empty_success_rate}
\end{table}

\section{Conclusion}
We presented the homotopic information gain for solving the active target tracking problem. Leveraging the homotopic belief first proposed in~\cite{wakulicz_topological_2023} we constructed a light-weight information gain measure comprising only discrete finite sums. In tandem with the new measure we proposed a planning approach that capitalises on the sparse nature of homotopic information gain. Our homotopic approach enables more efficient informative path planning than a conventional metric-information-based approach, as demonstrated in evaluation on real and simulated datasets.
Our evaluations revealed pathways to further improvements in future work, including investigation of different homotopy invariants and better handling of time-of-arrival uncertainty to improve success rate.


This work opens many exciting opportunities for extending the homotopic tracking framework. The sparse nature of our approach can be further leveraged in resource-constrained scenarios such as multi-target tracking, or multi-task problems. Moreover, the discrete nature of the homotopic belief can be exploited in planning for multi-robot systems, where robots could be assigned tasks based on likely future motion hypotheses. Real robot deployments for these tasks will require moving to a receding horizon planning formulation as the remaining length of target trajectories will be unknown. Then, with short-term predictions our proposed approach will enable re-planning every second.

\bibliographystyle{IEEEtran}
\bibliography{bib}

@book{matrixcookbook,
author = {Petersen, Kaare Breandt and Pedersen, Michael Syskind},
 
isbn = {0962-1083 (Print)$\backslash$r0962-1083 (Linking)},
 
journal = {Citeseer},
number = {4},
pages = {1--66},
pmid = {17284204},
title = {The Matrix Cookbook},
volume = {16},
year = {2007},
publisher = {Technical University of Denmark}
}

@article{atc_dataset,
    author = {Brscic, D. and Kanda, T. and Ikeda, T. and Miyashita, T.},
    title = {Person position and body direction tracking in large public spaces using 3{D} range sensors},
    journal = {Trans. on Human-Mach. Sys.},
    volume={43},
    issue = {6},
    pages = {522 - 534},
    year = {2013}
}

@book{hatcher_algebraic_2002,
	title = {Algebraic {Topology}},
	isbn = {978-0-521-79540-1},
	 
	publisher = {Cambridge University Press},
	author = {Hatcher, A.},
	year = {2002},
	lccn = {00065166},
}

@article{pokorny_topological_2016,
	title = {Topological trajectory classification with filtrations of simplicial complexes and persistent homology},
	volume = {35},
	number = {1-3},
	journal = {International Journal of Robotics Research},
	author = {Pokorny, Florian T. and Hawasly, Majd and Ramamoorthy, Subramanian},
	year = {2016},
	pages = {204--223},
}

@article{thorDataset2019,
  title={TH{\"O}R: Human-Robot Navigation Data Collection and Accurate Motion Trajectories Dataset},  author={Rudenko, Andrey and Kucner, Tomasz P and Swaminathan, Chittaranjan S and Chadalavada, Ravi T and Arras, Kai O and Lilienthal, Achim J},
  journal={IEEE Robotics and Automation Letters},
  volume={5},
  number={2},
  pages={676--682},
  year={2020},
  publisher={IEEE}
}

@article{mccammon_topological_2021,
	title = {Topological path planning for autonomous information gathering},
	 
	 
	 
	journal = {Autonomous Robots},
	author = {McCammon, Seth and Hollinger, Geoffrey A.},
	year = {2021},
}

@article{zyner_naturalistic_2020,
	title = {Naturalistic {Driver} {Intention} and {Path} {Prediction} {Using} {Recurrent} {Neural} {Networks}},
	volume = {21},
	 
	 
	 
	number = {4},
	 
	journal = {IEEE Transactions on Intelligent Transportation Systems},
	author = {Zyner, Alex and Worrall, Stewart and Nebot, Eduardo},
	  
	year = {2020},
	pages = {1584--1594},
}

@article{Ivanovic_2019_ICCV,
author = {Ivanovic, Boris and Pavone, Marco},
title = {The Trajectron: Probabilistic Multi-Agent Trajectory Modeling With Dynamic Spatiotemporal Graphs},
journal = {Proceedings of the IEEE/CVF International Conference on Computer Vision   },
 
year = {2019}
}

@article{zhi_kernel_2020,
	title = {Kernel {Trajectory} {Maps} for {Multi}-{Modal} {Probabilistic} {Motion} {Prediction}},
	volume = {100},
	journal = {Proceedings of the {Conference} on {Robot} {Learning}},
	author = {Zhi, Weiming and Ott, Lionel and Ramos, Fabio},
	  
	year = {2020},
	pages = {1405--1414},
}

@article{kiss_constrained_2022,
	title = {Constrained {Gaussian} {Processes} {With} {Integrated} {Kernels} for {Long}-{Horizon} {Prediction} of {Dense} {Pedestrian} {Crowd} {Flows}},
	volume = {7},
	 
	 
	 
	number = {3},
	 
	journal = {IEEE Robotics and Automation Letters},
	author = {Kiss, Stefan H. and Katuwandeniya, Kavindie and Alempijevic, Alen and Vidal-Calleja, Teresa},
	  
	year = {2022},
	pages = {7343--7350},
}

@article{wakulicz_active_2021,
	address = {Xi'an, China},
	title = {Active {Information} {Acquisition} under {Arbitrary} {Unknown} {Disturbances}},
	isbn = {978-1-72819-077-8},
	 
	 
	 
	journal = {{IEEE} {International} {Conference} on {Robotics} and {Automation}   },
	publisher = {IEEE},
	author = {Wakulicz, Jennifer and Kong, He and Sukkarieh, Salah},
	  
	year = {2021},
	pages = {8429--8435}}

@article{lavalle_sensing_2010,
	title = {Sensing and {Filtering}: {A} {Fresh} {Perspective} {Based} on {Preimages} and {Information} {Spaces}},
	volume = {1},
	 
	shorttitle = {Sensing and {Filtering}},
	 
	 
	language = {en},
	number = {4},
	 
	journal = {Foundations and Trends in Robotics},
	author = {LaValle, Steven M.},
	year = {2010},
	pages = {253--372},
}

@article{zheng_gaussian_2024,
	title = {A {Gaussian} mixture multiple-model belief propagation filter for multisensor-multitarget tracking},
	volume = {220},
	 
	 
	 
	language = {en},
	 
	journal = {Signal Processing},
	author = {Zheng, Feng and Tian, Yu and Zhan, Weicong and Yu, Jiancheng and Liu, Kaizhou},
	 
	year = {2024},
	pages = {109473},
}

@article{billard_nonmyopic_2023,
	  
	title = {Nonmyopic {Distilled} {Data} {Association} {Belief} {Space} {Planning} {Under} {Budget} {Constraints}},
	volume = {27},
	 
	 
	language = {en},
	 
	journal = {Robotics {Research}},
	 
	author = {Shienman, Moshe and Indelman, Vadim},
	year = {2023},
	 
	pages = {102--118}
}

@article{vo_gaussian_2006,
	title = {The {Gaussian} {Mixture} {Probability} {Hypothesis} {Density} {Filter}},
	volume = {54},
	 
	 
	 
	number = {11},
	 
	journal = {IEEE Transactions on Signal Processing},
	author = {Vo, B.-N. and Ma, W.-K.},
	 
	year = {2006},
	pages = {4091--4104}
}

@article{kim_multiple_2015,
	  
	title = {Multiple {Hypothesis} {Tracking} {Revisited}},
	 
	 
	 
	 
	journal = {{IEEE} {International} {Conference} on {Computer} {Vision}},
	 
	author = {Kim, Chanho and Li, Fuxin and Ciptadi, Arridhana and Rehg, James M.},
	 
	year = {2015},
	pages = {4696--4704},
}

@article{ruijie_he_efficient_2010,
	  
	title = {Efficient planning under uncertainty for a target-tracking micro-aerial vehicle},
	 
	 
	 
	 
	journal = {{IEEE} {International} {Conference} on {Robotics} and {Automation}},
	 
	author = {{Ruijie He} and Bachrach, Abraham and Roy, Nicholas},
	 
	year = {2010},
	pages = {1--8},
}

@article{vitus_closed_loop_2011,
	  
	title = {Closed-loop belief space planning for linear, {Gaussian} systems},
	 
	 
	 
	 
	journal = {{IEEE} {International} {Conference} on {Robotics} and {Automation}},
	 
	author = {Vitus, Michael P. and Tomlin, Claire J.},
	 
	year = {2011},
	pages = {2152--2159},
}

@article{kurniawati_partially_2022,
	title = {Partially {Observable} {Markov} {Decision} {Processes} and {Robotics}},
	volume = {5},
	number = {1},
	journal = {Annual Review of Control, Robotics, and Autonomous Systems},
	author = {Kurniawati, Hanna},
	year = {2022},
	pages = {253--277},
}

@article{Kocsis_bandit_2006,
	  
	title = {Bandit {Based} {Monte}-{Carlo} {Planning}},
	volume = {4212},
	 
	 
	 
	journal = {Machine {Learning}},
	 
	author = {Kocsis, Levente and Szepesvári, Csaba},
	year = {2006},
	 
	pages = {282--293},
}

@article{bhattacharya_search-based_2010,
	  
	title = {Search-based path planning with homotopy class constraints},
	 
	journal = {{AAAI} {Conference} on {Artificial} {Intelligence}},
	 
	author = {Bhattacharya, Subhrajit and Kumar, Vijay and Likhachev, Maxim},
	 
	year = {2010},
	pages = {1230--1237},
}

@article{frederico_carvalho_long-term_2019,
	  
	title = {Long-term {Prediction} of {Motion} {Trajectories} {Using} {Path} {Homology} {Clusters}},
	 
	 
	 
	 
	journal = {{IEEE}/{RSJ} {International} {Conference} on {Intelligent} {Robots} and {Systems}  },
	 
	author = {Frederico Carvalho, J. and Vejdemo-Johansson, Mikael and Pokorny, Florian T. and Kragic, Danica},
	 
	year = {2019},
	pages = {765--772},
}

@ARTICLE{eiffert_probabilistic_2020,

  author={Eiffert, Stuart and Li, Kunming and Shan, Mao and Worrall, Stewart and Sukkarieh, Salah and Nebot, Eduardo},

  journal={IEEE Robotics and Automation Letters}, 

  title={Probabilistic Crowd GAN: Multimodal Pedestrian Trajectory Prediction Using a Graph Vehicle-Pedestrian Attention Network}, 

  year={2020},

  volume={5},

  number={4},

  pages={5026-5033},
}

@article{li_attentional-gcnn_2020,
  author={Li, Kunming and Eiffert, Stuart and Shan, Mao and Gomez-Donoso, Francisco and Worrall, Stewart and Nebot, Eduardo},
  journal={IEEE International Conference on Robotics and Automation}, 
  title={Attentional-GCNN: Adaptive Pedestrian Trajectory Prediction towards Generic Autonomous Vehicle Use Cases}, 
  year={2021},
  volume={},
  number={},
  pages={14241-14247},
   }

@article{wiest_probabilistic_2012,
	  
	title = {Probabilistic trajectory prediction with {Gaussian} mixture models},
	 
	 
	 
	 
	journal = {{IEEE} {Intelligent} {Vehicles} {Symposium}},
	 
	author = {Wiest, Jurgen and Hoffken, Matthias and Kresel, Ulrich and Dietmayer, Klaus},
	 
	year = {2012},
	pages = {141--146},
}

@article{tian_lan_estimating_2006,
	  
	title = {Estimating {Mutual} {Information} {Using} {Gaussian} {Mixture} {Model} for {Feature} {Ranking} and {Selection}},
	 
	 
	 
	 
	journal = {{IEEE} {International} {Joint} {Conference} on {Neural} {Network} {Proceedings}},
	 
	author = {{Tian Lan} and Erdogmus, D. and Ozertem, U. and {Yonghong Huang}},
	year = {2006},
	pages = {5034--5039},
}

@article{siciliano_sensor_2009,
	  
	title = {Sensor {Beams}, {Obstacles}, and {Possible} {Paths}},
	volume = {57},
	 
	 
	 
	journal = {Algorithmic {Foundation} of {Robotics} {VIII}},
	 
	author = {Tovar, Benjamin and Cohen, Fred and LaValle, Steven M.},
	year = {2009},
	 
	pages = {317--332},
}

@article{carreira-perpinan_mode-finding_2000,
	title = {Mode-finding for mixtures of {Gaussian} distributions},
	volume = {22},
	 
	 
	 
	number = {11},
	 
	journal = {IEEE Transactions on Pattern Analysis and Machine Intelligence},
	author = {Carreira-Perpinan, M.A.},
	 
	year = {2000},
	pages = {1318--1323},
}

@article{hershey_approximating_2007,
	  
	title = {Approximating the {Kullback} {Leibler} {Divergence} {Between} {Gaussian} {Mixture} {Models}},
	 
	 
	 
	 
	journal = {{IEEE} {International} {Conference} on {Acoustics}, {Speech} and {Signal} {Processing}},
	 
	author = {Hershey, John R. and Olsen, Peder A.},
	 
	year = {2007},
	pages = {IV--317--IV--320},
}

@article{kolchinsky_estimating_2017,
	title = {Estimating {Mixture} {Entropy} with {Pairwise} {Distances}},
	volume = {19},
	 
	 
	 
	language = {en},
	number = {7},
	 
	journal = {Entropy},
	author = {Kolchinsky, Artemy and Tracey, Brendan},
	 
	year = {2017},
	pages = {361},
}

@article{kantor_orienteering_1992,
	title = {The {Orienteering} {Problem} with {Time} {Windows}},
	volume = {43},
	 
	 
	 
	number = {6},
	 
	journal = {The Journal of the Operational Research Society},
	author = {Kantor, Marisa G. and Rosenwein, Moshe B.},
	 
	year = {1992},
	pages = {629},
}

@INPROCEEDINGS{sakcak_convex_hsig,
  author={Sakcak, Basak and Bascetta, Luca and Ferretti, Gianni},
  booktitle={2019 18th European Control Conference (ECC)}, 
  title={Homotopy aware kinodynamic planning using RRT-based planners}, 
  year={2019},
  volume={},
  number={},
  pages={1568-1573},
}

@article{wakulicz_topological_2023,
	  
	title = {Topological {Trajectory} {Prediction} with {Homotopy} {Classes}},
	 
	 
	 
	 
	journal = {{IEEE} {International} {Conference} on {Robotics} and {Automation}  },
	 
	author = {Wakulicz, Jennifer and Brian Lee, Ki Myung and Vidal-Calleja, Teresa and Fitch, Robert},
	 
	year = {2023},
	pages = {6930--6936},
}

@article{pmlr-v119-fischer20a,
  title = 	 {Information Particle Filter Tree: An Online Algorithm for {POMDP}s with Belief-Based Rewards on Continuous Domains},
  author =       {Fischer, Johannes and Tas, \"Omer Sahin},
  journal = 	 {International Conference on Machine Learning},
  pages = 	 {3177--3187},
  year = 	 {2020},
  volume = 	 {119}
}

@article{ryan_particle_2010,
	title = {Particle filter based information-theoretic active sensing},
	volume = {58},
	 
	 
	 
	language = {en},
	number = {5},
	 
	journal = {Robotics and Autonomous Systems},
	author = {Ryan, Allison and Hedrick, J. Karl},
	 
	year = {2010},
	pages = {574--584},
}

@article{Pappas2009,
  author = {J. Le Ny and G. Pappas},
  title = {On trajectory optimization for active sensing in Gaussian process models},
  journal = {IEEE Conference on Decision and Control},
  pages = {6286--6292},
  year = {2009}
}

@article{Schlotfeldt2018,
  author = {B. Schlotfeldt and D. Thakur and N. Atanasov and V. Kumar and G. Pappas},
  title = {Anytime planning for decentralized multirobot active information gathering},
  journal = {IEEE Robotics and Automation Letters},
  volume = {3},
  number = {2},
  pages = {1025--1032},
  year = {2018}
}

@article{Tokekar2016,
  author = {Z. Zhang and P. Tokekar},
  title = {Non-myopic target tracking strategies for non-linear systems},
  journal = {IEEE Conference on Decision and Control},
  pages = {5591--5596},
  year = {2016}
}

@article{kantaros2019,
  author = {Y. Kantaros and B. Schlotfeldt and N. Atanasov and G. Pappas},
  title = {Asymptotically optimal planning for non-myopic multi-robot information gathering},
  journal = {Robotics: Science and Systems},
  pages = {22--26},
  year = {2019}
}

@article{Tomlin2012,
  author = {M. Vitus and W. Zhang and A. Abate and J. Hu and C. Tomlin},
  title = {On efficient sensor scheduling for linear dynamical systems},
  journal = {Automatica},
  volume = {48},
  number = {10},
  pages = {2482--2493},
  year = {2012}
}

@article{Gillijns2007A,
	title = {Unbiased minimum-variance input and state estimation for linear discrete-time systems},
	volume = {43},
	copyright = {https://www.elsevier.com/tdm/userlicense/1.0/},
	 
	 
	 
	language = {en},
	number = {1},
	 
	journal = {Automatica},
	author = {Gillijns, Steven and De Moor, Bart},
	 
	year = {2007},
	pages = {111--116},
}

@article{zahroof_multi-robot_2023,
	  
	title = {Multi-{Robot} {Localization} and {Target} {Tracking} with {Connectivity} {Maintenance} and {Collision} {Avoidance}},
	 
	 
	 
	 
	journal = {{American} {Control} {Conference} ({ACC})},
	 
	author = {Zahroof, Rahul and Liu, Jiazhen and Zhou, Lifeng and Kumar, Vijay},
	 
	year = {2023},
	pages = {1331--1338}
}

@article{tokekar_multi-target_2014,
	  
	title = {Multi-target visual tracking with aerial robots},
	 
	 
	 
	 
	journal = {{IEEE}/{RSJ} {International} {Conference} on {Intelligent} {Robots} and {Systems}  },
	 
	author = {Tokekar, Pratap and Isler, Volkan and Franchi, Antonio},
	 
	year = {2014},
	pages = {3067--3072}
}

@article{atanasov_information_2014,
	title = {Information acquisition with sensing robots: {Algorithms} and error bounds},
	journal = {{IEEE} {International} {Conference} on {Robotics} and {Automation}},
	author = {Atanasov, N. and Ny, J. Le and Daniilidis, K. and Pappas, G. J.},
	year = {2014},
	pages = {6447--6454},
}

@article{tzoumas_sensor_2016,
	  
	title = {Sensor placement for optimal {Kalman} filtering: {Fundamental} limits, submodularity, and algorithms},
	 
	shorttitle = {Sensor placement for optimal {Kalman} filtering},
	 
	 
	 
	journal = {{American} {Control} {Conference} ({ACC})},
	 
	author = {Tzoumas, V. and Jadbabaie, A. and Pappas, G. J.},
	 
	year = {2016},
	pages = {191--196},
}

@article{zhou_resilient_2019,
	title = {Resilient {Active} {Target} {Tracking} {With} {Multiple} {Robots}},
	volume = {4},
	 
	 
	 
	number = {1},
	 
	journal = {IEEE Robotics and Automation Letters},
	author = {Zhou, Lifeng and Tzoumas, Vasileios and Pappas, George J. and Tokekar, Pratap},
	  
	year = {2019},
	pages = {129--136},
}

@article{zhou_active_2019,
	title = {Active {Target} {Tracking} {With} {Self}-{Triggered} {Communications} in {Multi}-{Robot} {Teams}},
	volume = {16},
	 
	 
	 
	number = {3},
	 
	journal = {IEEE Transactions on Automation Science and Engineering},
	author = {Zhou, Lifeng and Tokekar, Pratap},
	 
	year = {2019},
	pages = {1085--1096},
}

@article{ke_zhou_multirobot_2011,
	title = {Multirobot {Active} {Target} {Tracking} {With} {Combinations} of {Relative} {Observations}},
	volume = {27},
	 
	 
	 
	number = {4},
	 
	journal = {IEEE Transactions on Robotics},
	author = {{Ke Zhou} and Roumeliotis, S. I.},
	  
	year = {2011},
	pages = {678--695},
}

@article{de_groot_topology-driven_2025,
	title = {Topology-{Driven} {Parallel} {Trajectory} {Optimization} in {Dynamic} {Environments}},
	volume = {41},
	copyright = {https://ieeexplore.ieee.org/Xplorehelp/downloads/license-information/IEEE.html},
	 
	 
	 
	 
	journal = {IEEE Transactions on Robotics},
	author = {De Groot, Oscar and Ferranti, Laura and Gavrila, Dariu M. and Alonso-Mora, Javier},
	year = {2024},
	pages = {110--126},
}

@article{yi_homotopy-aware_2016,
	address = {Christchurch, New Zealand},
	title = {Homotopy-aware {RRT}*: {Toward} human-robot topological path-planning},
	copyright = {https://doi.org/10.15223/policy-029},
	isbn = {978-1-4673-8370-7},
	shorttitle = {Homotopy-aware {RRT}*},
	 
	 
	 
	journal = {2016 11th {ACM}/{IEEE} {International} {Conference} on {Human}-{Robot} {Interaction} ({HRI})},
	publisher = {IEEE},
	author = {Yi, Daqing and Goodrich, Michael A. and Seppi, Kevin D.},
	  
	year = {2016},
	pages = {279--286},
}

@article{kolur_online_2019,
	title = {Online {Motion} {Planning} {Over} {Multiple} {Homotopy} {Classes} with {Gaussian} {Process} {Inference}},
	copyright = {arXiv.org perpetual, non-exclusive license},
	 
	author = {Kolur, Keshav and Chintalapudi, Sahit and Boots, Byron and Mukadam, Mustafa},
	year = {2019},
        journal = {IEEE/RSJ International Conference on Intelligent Robots and Systems},
pages={2358--2364},
}

@article{novosad_ctopprm_2023,
	title = {{CTopPRM}: {Clustering} {Topological} {PRM} for {Planning} {Multiple} {Distinct} {Paths} in {3D} {Environments}},
	volume = {8},
	copyright = {https://ieeexplore.ieee.org/Xplorehelp/downloads/license-information/IEEE.html},
	 
	shorttitle = {{CTopPRM}},
	 
	 
	number = {11},
	 
	journal = {IEEE Robotics and Automation Letters},
	author = {Novosad, Matej and Penicka, Robert and Vonasek, Vojtech},
	  
	year = {2023},
	pages = {7336--7343},
}

@book{el_gamal_network_2011,
	edition = {1},
	title = {Network {Information} {Theory}:},
	copyright = {https://www.cambridge.org/core/terms},
	publisher = {Cambridge University Press},
	author = {El Gamal, Abbas and Kim, Young-Han},
	  
	year = {2011},
}

@article{karamouzas17,
 author = {Karamouzas, Ioannis and Sohre, Nick and Narain, Rahul and Guy, Stephen J.},
 title = {Implicit Crowds: Optimization Integrator for Robust Crowd Simulation},
 journal = {ACM Transactions on Graphics},
 volume = {36},
 number = {4},
 year = {2017},
   
  
}

@article{pmlr-v211-yang23a,
  title = 	 {Policy Learning for Active Target Tracking over Continuous $SE(3)$ Trajectories},
  author =       {Yang, Pengzhi and Koga, Shumon and Asgharivaskasi, Arash and Atanasov, Nikolay},
  journal = 	 {Proceedings of The 5th Annual Learning for Dynamics and Control Conference},
  pages = 	 {64--75},
  year = 	 {2023},
  volume = 	 {211}
}

\newpage
 
\begin{IEEEbiography}[{\includegraphics[width=1in,height=1.25in,clip,keepaspectratio]{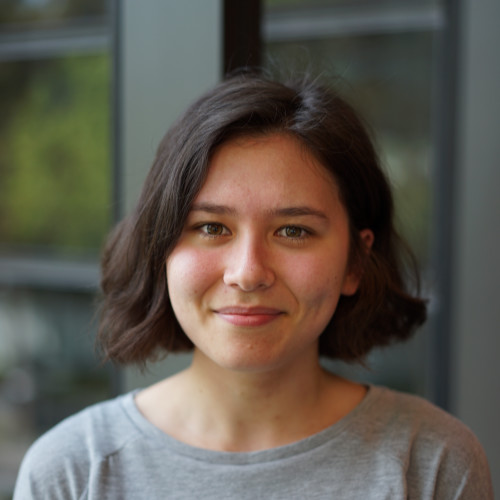}}]{Jennifer Wakulicz}
(Member, IEEE) is a Postdoctoral Research Fellow at the University of Sydney, Sydney, NSW, Australia.
She received the B.Sc. degree in Mathematics and Physics from the University of Sydney, Camperdown, NSW, Australia, and the Ph.D. degree in robotics from the University of Technology Sydney, Ultimo, NSW, Australia, in 2025.
Her research lies in probabilistic modelling and planning, with the objective of developing robust robotic decision-making frameworks capable of operating reliably in unstructured environments.
\end{IEEEbiography}

\begin{IEEEbiography}[{\includegraphics[width=1in,height=1.25in,clip,keepaspectratio]{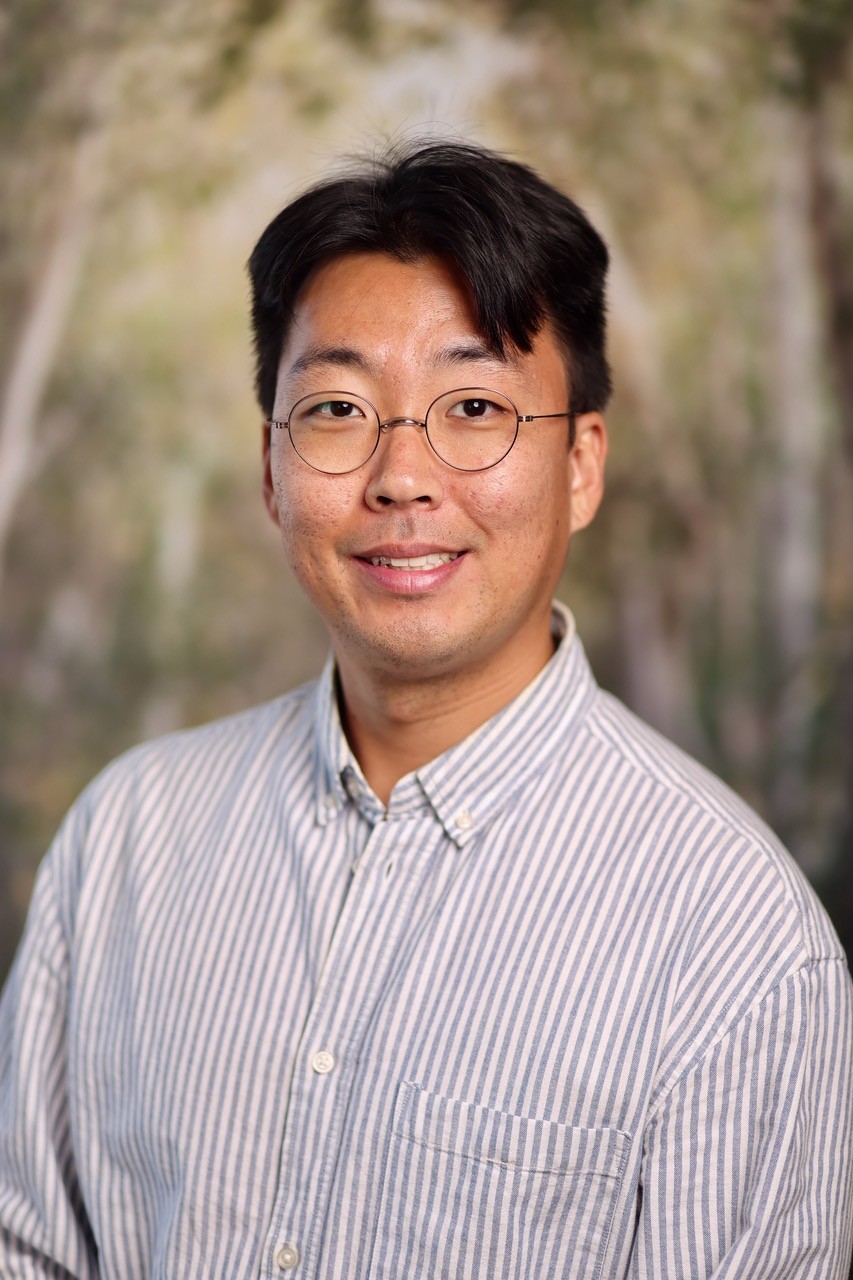}}]{Ki Myung Brian Lee}
    (Member, IEEE) is a Postdoctoral Scholar at the Contextual Robotics Institute, University of California San Diego, La Jolla, CA, USA. 
He received the B.Eng. (Hons I) degree in mechatronics (space) from the University of Sydney, Camperdown, NSW, Australia, and the Ph.D. degree in robotics from the University of Technology Sydney, Ultimo, NSW, Australia, in 2023.
He was recognized as an RSS Pioneer of 2023, and was the recipient of the UTS Research Excellence Scholarship. 
His current research aims to develop novel representations of environments and tasks that accelerate planning and control. More broadly, he is interested in mobile robot autonomy in previously unseen environments.
\end{IEEEbiography}

\begin{IEEEbiography}[{\includegraphics[width=1in,height=1.25in,clip,keepaspectratio]{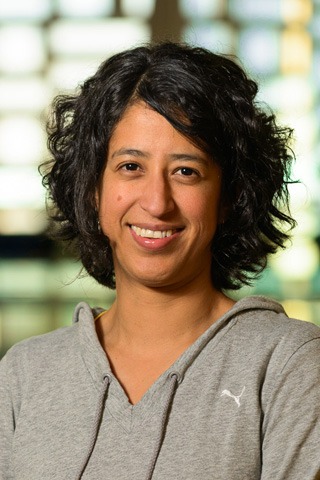}}]{Teresa Vidal-Calleja}
    (Senior Member, IEEE) received the B.Eng. degree in mechanical engineering from the National Autonomous University of Mexico, Mexico City, Mexico, in 2000, the M.Sc. degree in mechatronics from CINVESTAV-IPN, Mexico City, Mexico, in 2002, and the Ph.D. degree in automatic control, computer vision, and robotics from the Polytechnic University of Catalonia, Barcelona, Spain, in 2007. She is currently a Professor and Research Director of the Robotics Institute at the University of Technology Sydney (UTS), Sydney, Australia. She was previously a Postdoctoral Research Fellow with LAAS-CNRS, Toulouse, France, and with the Australian Centre for Field Robotics at the University of Sydney, Sydney, Australia. In 2012, she was appointed as a Chancellor’s Research Fellow with the Robotics Institute at UTS. She has also been a Visiting Professor at the Tokyo University of Science, Tokyo, Japan, and at the Institute of Robotics and Mechatronics, German Aerospace Centre (DLR), Cologne, Germany, as well as a Visiting Scholar with the Active Vision Laboratory, University of Oxford, Oxford, U.K., and the Autonomous Systems Lab, ETH Zürich, Zürich, Switzerland. Her research focuses on robotic perception, combining estimation theory and machine learning.
\end{IEEEbiography}

\begin{IEEEbiography}[{\includegraphics[width=1in,height=1.25in,clip,keepaspectratio]{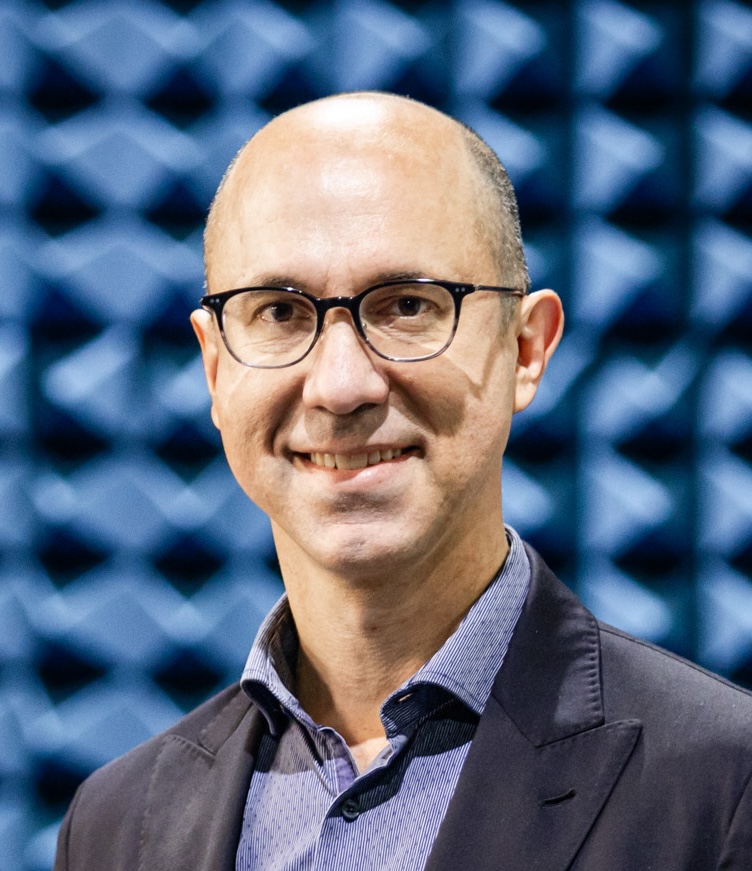}}]{Robert Fitch}
    (Member, IEEE) received the Ph.D. degree in computer science from Dartmouth College, USA, in 2005. He is currently a Professor in the School of Mechanical and Mechatronic Engineering at the University of Technology Sydney (UTS), Australia. His research interests include motion planning, field robotics, multi-robot systems, and coordinated autonomy.
\end{IEEEbiography}

\end{document}